\documentclass[fleqn,12pt]{olplainarticle}
\title{A Meta-heuristic Approach to Estimate and Explain Classifier Uncertainty}

\author[1,2]{Andrew Houston}
\author[1]{Georgina Cosma}
\affil[1]{Department of Computer Science, Loughborough University, Epinal Way, Loughborough, UK}
\affil[2]{Academic Department of Military Rehabilitation, Defence Medical Rehabilitation Centre, Stanford Hall, Loughborough, UK}


\begin{abstract}
Trust is a crucial factor affecting the adoption of machine learning (ML) models. Qualitative studies have revealed that end-users, particularly in the medical domain, need models that can express their uncertainty in decision-making allowing users to know when to ignore the model's recommendations. However, existing approaches for quantifying decision-making uncertainty are not model-agnostic, or they rely on complex statistical derivations that are not easily understood by laypersons or end-users, making them less useful for explaining the model's decision-making process. This work proposes a set of class-independent meta-heuristics that can characterize the complexity of an instance in terms of factors are mutually relevant to both human and ML decision-making. The measures are integrated into a meta-learning framework that estimates the risk of misclassification. The proposed framework outperformed predicted probabilities in identifying instances at risk of being misclassified. The proposed measures and framework hold promise for improving model development for more complex instances, as well as providing a new means of model abstention and explanation.
\end{abstract}

\begin{document}

\flushbottom
\maketitle
\thispagestyle{empty}

\section{Introduction}
\noindent In recent years, significant advancements have been made in the application of machine learning (ML) to support clinical decision-making processes. The medical domain has seen various applications of ML, including the prediction of surgical outcomes~\cite{karhade2019development,polce2020development,houston2021predicting} to aid in treatment planning, earlier diagnoses of cancers~\cite{kourou2015machine, mckinney2020international} to improve patient survival rates, and prognostic tools for better management of neuro-degenerative conditions~\cite{dallora2017machine, patel2021artificial}. Despite their potential benefits, the adoption of such tools remains a challenge, with trust being cited as a primary barrier~\cite{asan2020artificial}. Complex, ‘black-box’, algorithms that cannot explain when they may be in correct are often cited a source of distrust~\cite{asan2020artificial}. As the use of AI continues to grow in fields where incorrect decisions can have serious consequences, there is a growing need to equip end-users with tools to facilitate an appropriate trust relationship with ML and AI tools, termed trust calibration.

To address the demands for more interpretable models several approaches have been proposed, including the use of interpretability and explainability measures such as LIME~\cite{ribeiro2016should} and SHAP~\cite{lundberg2017unified}, which offer post-hoc explanations for model predictions. Another technique is the utilisation of inherently interpretable models, such as decision trees~\cite{quinlan1986induction}. However, as highlighted by Kaur et al.~\cite{kaur2022sensible}, these methods may not take into account the contextual factors that influence how end-users internalise information. Moreover, in practice, even explanations of accurate model predictions have been shown to be insufficient to overcome the internal biases of incorrect end-users~\cite{kohli2018cad,cole2014impact,xie2020chexplain}.

The design of methods to facilitate trust calibration requires consideration of the needs and perspectives of the end user. In the medical domain, clinicians have emphasized the importance of models indicating uncertainty or abstaining when their confidence is low~\cite{kompa2021second,tonekaboni2019clinicians}. Global measures of model performance, while useful for gauging overall reliability, are not enough to foster trust and sustained use, as they lack insight into individual cases~\cite{tonekaboni2019clinicians,rechkemmer2022confidence}. Therefore, it is crucial to develop methods that are interpretable and applicable at the instance level. Amann et al.~\cite{amann2020explainability} emphasise that the effectiveness of explanations depends on the end user's ability to comprehend their meaning. Thus, the interpretability of the methods themselves is a crucial factor in facilitating trust calibration.

This paper presents a novel approach to calibrating trust in machine learning (ML) models, incorporating the following three key contributions:
\begin{itemize}
    \item A suite of model-agnostic, interpretable meta-heuristics that aim to characterise instances in terms of the sources of complexity in decision-making.
    \item A meta-learning framework, incorporating a synthetic data generator and Bayesian-optimized, weighted fuzzy-clustering system to estimate the level of uncertainty in the decision-making of an ML model, that outperforms predictive probabilities for characterising misclassification risk.
    \item Experiments evaluating how the proposed methods could enable ML models to refrain from making predictions when uncertainty is high and how uncertainty can be communicated using information derived from the meta-features.
\end{itemize}

The remainder of paper is structured as follows: Section~\ref{sec:Related_Works} provides a descriptive overview of sources of uncertainty in decision making and existing approaches to characterising the complexity of instances and uncertainty of decision made by ML models;  Section~\ref{sec:Proposed_Methods} describes the proposed meta-heuristics, outlines the design of the fuzzy clustering system for estimating uncertainty and describes the design of a knowledge-base used to improve the performance of the uncertainty estimation system; Section~\ref{sec:Experimental_Methods} details the methods for analysing the relationships between the proposed meta-heuristics and misclassification events, identifying the optimal method for knowledge-base construction, and evaluating the performance of the uncertainty estimation system; Section~\ref{sec:Experimental_Results} provides the experimental results; Section~\ref{sec:Applications} explores the use of the proposed methods for abstention and uncertainty explanation; Lastly, section~\ref{sec:Discussion} provides a discussion of future directions.

\section{Related Works}\label{sec:Related_Works}
\noindent This section provides a descriptive overview of the types of uncertainty in decision making, data-centric sources of complexity in machine learning tasks and existing methods to characterise such sources of complexity and estimate the uncertainty of decisions made by ML models.

\subsection{Types of Uncertainty}
\noindent Uncertainty can be classed as one of two types, Epistemic or aleatoric. \textit{Epistemic uncertainty} can be described as a type of uncertainty originating from the insufficiency of similar training data~\cite{hullermeier2021aleatoric}. Epistemic uncertainty can manifest in various forms, such as the absence of underrepresented groups in facial recognition datasets, resulting in a decline in recognition performance for these groups~\cite{buolamwini2017gender}, or in the occurrence of rare circumstances within a dataset. \textit{Aleatoric uncertainty} reflects uncertainty arising from a degree of randomness which cannot be explained away, such as the roll of a dice, flip of a coin, noise in a signal or low resolution of an image~\cite{spiegelhalter2008understanding}.\newline
In machine learning research, several factors have been identified for increasing the epistemic and aleatoric uncertainty of classification problems, which class imbalance, class overlap and outliers, and these are described below.

\subsubsection*{Class Imbalance}
\noindent Class imbalance is defined as an unequal distribution of instances between classes in a dataset and is a common problem in many domains, spanning medical predictions~\cite{houston2021predicting}, sentiment analysis~\cite{ghosh2019imbalanced} and information retrieval~\cite{chang2003statistical}. Large class imbalances can result in models that are highly accurate, but lack sensitivity~\cite{lopez2013insight}. Common approaches for addressing class imbalance include re-sampling techniques, such as SMOTE~\cite{chawla2002smote}, that either increase the instances in the minority class or reduce the majority class, and cost-sensitive learning, which gives higher weight to errors made on specific classes~\cite{elkan2001foundations}. However, class imbalance may have limited impact on the classifier performance, depending on other factors such as the well-defined class boundaries~\cite{vuttipittayamongkol2021class,smith2014instance}.

\subsubsection*{Class Overlap}
\noindent Class overlap, defined as the overlap in the feature space between instances of multiple classes, is a well-recognised challenge in classification tasks~\cite{vuttipittayamongkol2021class,smith2014instance}. The complexity of an instance in a region of class overlap is higher compared to instances in regions dominated by a single class. This overlap can introduce noise in a classification, increasing the aleatoric uncertainty of decisions made on such instances. The relationship between class overlap and class imbalance is often discussed in literature, with some studies suggesting that the impact of class overlap is greater than class imbalance, while the influence of class imbalance on the complexity of a classification task increases in problems with high overlap~\cite{vuttipittayamongkol2021class,smith2014instance,stefanowski2013overlapping}. In the context of clinical decision-making, class overlap may present in two forms: a general overlap of all features, making it difficult to distinguish between the diagnoses or prognoses of patients, or in patients where different aspects of their presentation align closely with different class outcomes. This is depicted in Fig.~\ref{fig:hypothetical dataset}, where instance D represents a patient with features $x1$ and $x2$ falling within the global overlap of the blue and red classes, and instance B represents a patient where feature $x1$ aligns closely with the blue class and feature $x2$ aligns with the red class.

\subsubsection*{Outliers}
\begin{figure}[ht]
    \centering
    \includegraphics[scale=0.7]{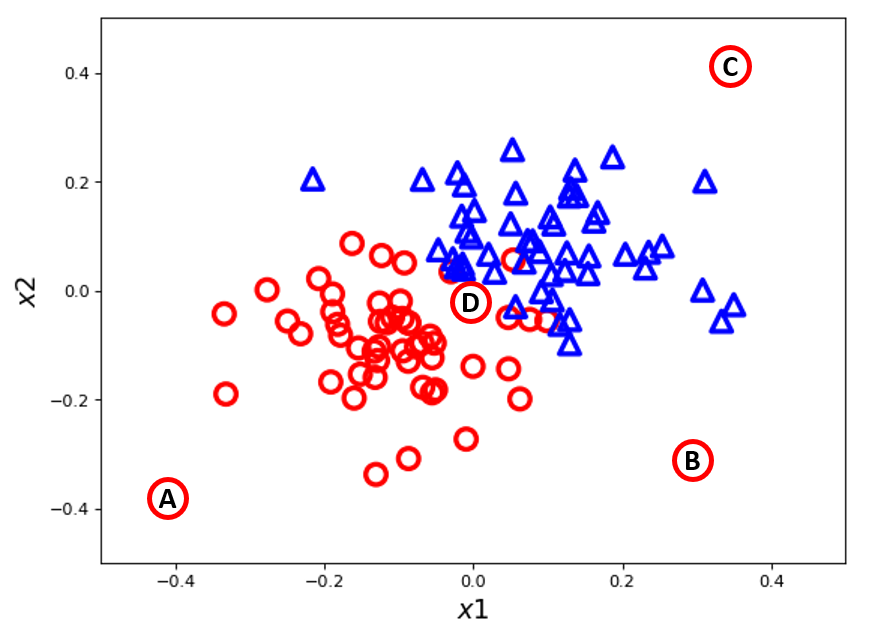}
    \caption{A hypothetical 2-feature dataset where points A and C are instances where a prediction could be considered to have a high degree of epistemic uncertainty, point B reflects an instance where a prediction could be considered to have both high epistemic and aleatoric uncertainty, and point D is an instance where a prediction could be considered to have a high degree of aleatoric uncertainty }
    \label{fig:hypothetical dataset}
\end{figure}

\noindent Outliers are instances that lie in a region of low neighborhood density and can result in higher levels of epistemic uncertainty in predictions due to the absence of similar instances for comparison. The impact of outliers on the complexity of a prediction is dependent on various factors, such as the location of the instance in the feature space and the classifier used. For instance, in the hypothetical dataset shown in Fig.\ref{fig:hypothetical dataset}, two instances may have similar levels of outlierness but their location in the feature space may result in differing levels of complexity. While instance A has high levels of outlierness, its location in the feature space relative to other instances of the same class may result in highly accurate predictions. On the other hand, instance C, though having similar levels of outlierness, may not be predicted accurately as its location in the feature space aligns closely with the opposing class. The impact of outliers on decision-making complexity is therefore specific to the classifier and the dataset \cite{acuna2005empirical}. In clinical decision-making, the availability of past evidence and experience plays a critical role in supporting predictions. Qualitative work found that outlying, abstract patient presentations can increase uncertainty in decision-making~\cite{islam2014heuristics}. Furthermore, due to the core underpinning of clinical decision-making being the availability of evidence and past experience~\cite{cioffi2001study,yang2019unremarkable}, in the case of rare and complex patient presentations, much like the presence of outliers within the context of ML, uncertainty in the decisions made increases~\cite{taruscio2021multifactorial}.

\subsection{Existing Approaches for Characterising Complexity and Uncertainty}

\subsubsection*{Meta-heuristic Approaches}
\noindent In 2011, Smith and Martinez~\cite{smith2011improving} proposed a series of hardness measures to identify instances that are likely to be misclassified in typical classification problems and defined thresholds for their use as a means of data cleaning during model development. A subsequent paper further explored the application of the proposed measures and their relevance to the complexity of instance-level decision-making~\cite{smith2014instance} finding class overlap to be most significant factor in characterizing complex instances. The paper presented several approaches for integrating the proposed hardness measures into the learning process, including the augmentation of error functions within multi-layer perceptrons to reduce the weight assigned to more complex instances and the removal of complex instances to reduce unwanted noise within the training set. The advantage of the measures proposed by Smith et al.~\cite{smith2014instance} is that they allow for the accurate characterization of the complexity of individual instances, enabling robust model development and improved reasoning as to why misclassifications occur, as demonstrated in  Houston et al.~\cite{houston2021predicting}. Additionally, the measures are easy to calculate, and the sources of complexity they reflect are well-defined, making them easily understood by end-users. However, post-deployment, most of the proposed measures are limited in their utility because they rely on class knowledge, which is unavailable in prospective cases.
 
In 2022, Barandas et al.~\cite{barandas2022uncertainty} proposed a method for characterizing aleatoric and epistemic uncertainty in traditional classification problems and applied the methods to aid in abstaining from making a prediction. They applied three measures to characterize the domains of uncertainty. Entropy was used to measure aleatoric uncertainty, variation ratios were applied to evaluate epistemic uncertainty resulting from the model, and a density estimation technique was applied to measure epistemic uncertainty resulting from a lack of data. The methods successful in utilizing uncertainty measures to improve the performance of a series of classifiers using uncertainty-based rejection. The authors augmented the way two models can be compared by looking at both actual performance metrics, such as accuracy, and the uncertainty values associated with the predictions. Additionally, the interpretable nature of the proposed measurements shows promise in acting as a facilitator of trust to the end user. However, a primary limiting factor for the application of such methods is their dependence on the classifier and its parameters. The difficulty with having model-dependent methods for characterizing instances is their limitation for use within meta-learning solutions, such as developing dynamic models and ensembles which, in some cases, relies on meta-information about an instance \cite{cruz2015meta, lamari2021smote}.

\subsubsection*{Model-Dependent Approaches}
\noindent The most straightforward way to quantify uncertainty is to use the prediction probability of the predicted class for the instance. Typically, prediction probability refers to the conditional probability of the predicted class, although it may differ depending on the model. For example, for SVM models, Platt scaling is commonly used to compute the prediction probability~\cite{platt1999probabilistic}. Rechkemmer and Yin~\cite{rechkemmer2022confidence} investigated the impact of predicted probabilities on end-users' decision-making and found that higher predicted probabilities increased users' willingness to follow the model's predictions and improved their self-reported trust in the model. This simple method for quantifying uncertainty has the advantage of being applicable to new instances post-deployment. However, it lacks interpretability as it does not explain why a model is less confident.

Active learning is an approach designed to reduce the computational cost and time required to train a learning algorithm by identifying the most useful instances to train a classifier. A popular method of active learning is uncertainty sampling, introduced by Lewis
181 and Gale~\cite{lewis1994sequential}, which identifies instances that a model is most uncertain about, learning the representation of such challenging cases to create a decision boundary. However, while uncertainty sampling is effective in identifying instances that a model is most uncertain about, like prediction probabilities, it does not identify the reasoning behind the uncertainty. Sharma
and Bilgic~\cite{sharma2017evidence} proposed an evidence-based framework for explaining uncertainty, targeting two specific sources of uncertainty which they termed `conflicting evidence uncertainty' and `insufficient evidence uncertainty'. `Conflicting evidence uncertainty' refers to the presence of strong evidence for an instance belonging to more than one class, whereas `insufficient evidence uncertainty' refers to a lack of evidence for an instance belonging to any class. Their evaluations on real-world datasets found that `conflicting evidence uncertainty' appeared to be a more effective means of active learning, outperforming traditional active learning and `insufficient evidence uncertainty'. The benefit of using such measures, unlike the hardness measures proposed by Smith et al.~\cite{smith2014instance}, is their lack of reliance on class knowledge. However, like the methods proposed in Barandas et al.~\cite{barandas2022uncertainty}, a downside to uncertainty measures derived through active learning is their lack of independence from a classifier. Additionally, active learning approaches typically fail to capture concepts such as noise and outlierness~\cite{roy2001toward}.

In 2021, Northcutt et al.~\cite{northcutt2021confident} proposed the confident learning (CL) algorithm to identify uncertainty in dataset labels and to detect instances that are incorrectly labelled. The CL algorithm estimates the joint distribution of noisy labels and latent labels by utilizing the out-of-sample predicted probabilities and noisy labels. This is done to characterize the class-conditional label noise and identify instances with labelling issues. These instances are then pruned, and the model is trained with re-weighting instances using the estimated latent priors. Applying CL has the major benefit of making a model more robust against epistemic error. Recently, Abad and Lee~\cite{abad2021detecting} utilized the CL algorithm to identify instances for which classifications were uncertain in a study focusing on mortality prediction. The study trained six ML models on a dataset, and then the CL algorithm was used to identify instances where class labels were uncertain. An XGBoost model was then retrained, predicting three classes, including a ``challenging'' patient class. Results were mixed, with the model achieving only a 31\% precision and 14\% recall for the ``challenging'' instances. However, after the uncertain patients were separately labelled, the XGBoost model's performance for identifying mortality patients increased from 87\% to 96\%. The work of Abad and Lee~\cite{abad2021detecting} demonstrates how the application of uncertainty identification can be used to improve classification performance and notify end-users of uncertain predictions. Despite the somewhat positive results, the application of the CL algorithm to identify uncertainty may be challenging to explain to end-users, similar to the use of conditional probabilities. This limits its applicability for explainability purposes, leaving the reasoning behind the uncertainty relatively unknown.

\subsubsection*{Summary}
\noindent To summarise, current methods for characterizing instance complexity and quantifying decision-making uncertainty have limitations, including post-deployment applicability, failure to consider multiple sources of complexity, and a lack of suitability for end-users to understand why a decision is uncertain or an instance is complex. Therefore, the proposed methods for characterising uncertainty take a user-focused approach to characterise complexity and estimate uncertainty, developing heuristics which reflect the factors that impact a human-made decision.

\subsection{Characterising Class Diversity}
\noindent The hardness measures of Smith et al.~\cite{smith2014instance} rely on class knowledge, often looking at class disagreement with the new instance, which limits their use post-deployment. Therefore, a class-independent alternative to disagreement is required. This study proposes the use of a diversity measure, observing how diverse the classes of the returned instances are as a measure of complexity. The calculation of diversity can be thought of in a similar way to calculating class imbalance.

Several methods have been proposed to quantify class imbalance in binary and multi-class problems, the most informative measures being empirical distributions and frequencies of classes within a dataset. However, these measures can be difficult to examine in highly multi-class problems and are not single-value representations. One proposed solution, suitable for multi-class problems is the imbalance ratio, proposed by Prati et al.~\cite{prati2015class}, providing a single-value summary of class imbalance. The imbalance ratio measures the number of instances in the most probable class for every instance in the least probable class. However, despite being effective in binary problems, the ratio is incapable of describing the class disparity in multi-class solutions due to its disregard for all classes other than the most and least probable. In recognising this flaw, Ortigosa-Hern\a'ndez et al.~\cite{ortigosa2017measuring} proposed the imbalance degree, a measure capable of characterising skewed class distributions in multi-class problems, using a similarity/distance function. The limitation of their approach was that the measure was sensitive to the choice of distance metric, which was prone to change across different classification problems. Acknowledging this,  Zhu et al.~\cite{zhu2018lrid} proposed a multi-class imbalanced degree, based on the likelihood-ratio test (LRID). Empirical evaluations on real and synthetic data sets showed their approach to be superior to the imbalance ratio and imbalance degree, with the additional benefit of not having to identify appropriate parameters for each dataset. Considering this study aims to propose generalisable methods, capable of identify complex instances across a range of classification tasks, the likelihood ratio imbalance degree will be applied to reflect the diversity within two of our proposed meta-heuristics, characterising diversity as:

\begin{equation}\label{Eq:Diversity Degree}
    Diversity = -2\sum_{c=1}^{C}m_cln\frac{b_c}{\hat{p}_c}
\end{equation}

\noindent where $C$ is the number of classes in the dataset, $m$ is the number of instances in each respective class, $ln$ is the natural logarithm, $b$ represents a balanced class distribution and $\hat{p}$ represents the estimated class distribution. $\hat{p}$ is estimated using:

\begin{equation}
    \hat{p} = \frac{m_c}{M}
\end{equation}

\noindent where $M$ is the total number of instances.

The diversity score is normalised to the worst possible imbalance within a dataset, for example: if a sample consists of 300 records and 3 classes, then the worst possible imbalance would be if a single class contained all 300 records. Contextualising this within the problem of estimating diversity, a normalised LRID of 1, would reflect zero diversity being present within our sample, as only one class is present.

\section{Proposed Meta-heuristics and Framework for Uncertainty Estimation}\label{sec:Proposed_Methods}
\noindent This section presents a series of meta-heuristics used to characterise the uncertainty of a model's prediction on a given instance and outlines the proposed framework for estimating the uncertainty of a prediction made by an ML model. 

\subsection{Meta-Heuristics for Characterising Instance Complexity}\label{subsec:meta-heuristics}
Seven meta-heuristics are proposed each requiring a dataset, $X$, containing $M$ instances and $N$ features, and an instance for which to characterise the complexity of, $x$.

\subsubsection*{$k$-Diverse Neighbours}
\noindent $k$-Diverse Neighbours ($k$DN) reflects the local overlap of an instance within the task space, relative to its nearest neighbours from the training set and is calculated according to Eq.~\ref{eq:KDN}, as:

\begin{equation}\label{eq:KDN}
    KDN(x) = diversity(KNN(x))
\end{equation}

\noindent where the function $KNN(x)$ returns the class labels of nearest $k$ instances in $X$ to $x$. 

\subsection*{Disjunct Size}
\noindent Disjunct size is a class-independent measure proposed by Smith et al.~\cite{smith2014instance} and calculates the size of the disjunct an instance is classified into by an unpruned decision tree, formed from a training set. The disjunct size of the returned instance is then normalised by dividing it by the size of the largest disjunct within the dataset. Disjunct size is calculated according to Eq.~\ref{eq:DS}, as:

\begin{equation}\label{eq:DS}
DS(x) = \frac{|distjunct(x)|-1}{\max_{d \in D}|disjunct(d)|-1}
\end{equation}

\noindent where $D$ is the set of all disjuncts in $X$, and $|distjunct(x)|-1$ returns size of the disjunct for instance $x$.

\subsubsection*{Disjunct Class Diversity}
\noindent Disjunct class diversity (DCD) reflects the diversity of the class of instances within the disjunct which an instance is classified into, ie those that are similar based on a subset of their features. Contrary to the methods of DS, the decision tree applied when calculating DCD is pruned. DCD is calculated according to Eq.~\ref{eq:DCD}, as:

\begin{equation}\label{eq:DCD}
    DCD(x) = diversity(Disjunt(x))
\end{equation}

\noindent where the function $Disjunt(x)$ returns the classes of the instances contained from dataset, $X$, in the same disjunct as instance, $x$.

\subsubsection*{Outlierness}
\noindent Outlierness (OL), reflects the degree to which an instance is similar to the instances contained within the training set. The outlierness of an instance is calculated using the density-based approach proposed by  Tang and He~\cite{tang2017local}, where the metric captures the ratio of the average neighbourhood density, to the density of instance $x$, using Eq.~\ref{eq:RDOS}, as:

\begin{equation}\label{eq:RDOS}
    OL(x) = \frac{\sum_{x\in S(x)}p(x)}{|S(x)|p(x)}
\end{equation}

\noindent where $S(x)$ is a set of instances formed of the $k$-nearest neighbours, $S_{KNN}(x)$, reverse nearest neighbours, $S_{RNN}(x)$, and shared nearest neighbours, $S_{SNN}(x)$, to instance $x$, termed a neighbourhood. The function $p(x)$ returns the density of the location of instance $x$, where density is calculated using Eq.~\ref{eq:density}, as:

\begin{equation}\label{eq:density}
    p(x) = \frac{1}{|S(x)|+1}\sum_{X\in S(x)\cup\{x\}}\frac{1}{h^N}K(\frac{X-x}{h})
\end{equation}

\noindent where $K\frac{X-x}{h}$ is a Gaussian kernel function with the kernel width of $h$ and neighbourhood size of $k$, $N$ is the number of features in $X$ and $|S(x)|$ is the size of the neighbourhood for instance, $x$.

\subsubsection*{Class-Level Outlierness}
\noindent Class-Level Outlierness (CL-OL), reflects the disparity in the level outlierness calculated for an instance, across each class within the dataset. Given an instance, an outlierness score is calculated for each class in the dataset. Thereafter, disparity is calculated using a variation of the diversity Eq.~\ref{Eq:Diversity Degree} where $m$, originally reflecting the number of instances in a class, now reflects the percentage of the summed outlierness scores, across all classes,$M$.

\subsubsection*{Evidence Conflict}
\noindent Evidence conflict (EC) reflects degree to which an instance's features fall within the distribution of a multiple classes. To calculate EC, first, an $M \times N$ matrix is formed, where $M$ refers to the number of classes within the dataset and $N$ refers to the number of features. The matrix will henceforth be referred to as the conflict matrix. To populate the conflict matrix, the conflicting evidence degree is calculated for each feature and class-comparison. The conflicting evidence degree (CED) is calculated using Eq.~\ref{eq:CED}, as:

\begin{equation}\label{eq:CED}
    CED(x^n) =     
    \begin{cases}
        1 - 1/(f(x_n, X_n^c)/f(x_n, X_n^r)) & f(x_n, X_n^c) > f(x_n, X_n^r)\\
        -1 + 1/(f(x_n, X_n^r)/f(x_n, X_n^c)) & f(x_n, X_n^c) > f(x_n, X_n^r)
    \end{cases}
\end{equation}

\noindent where the function $f(x_n, X_n^c)$ returns the probability density of the class $c$ for feature $n$ at the value of $x_n$ and the function $f(x_n, X_n^r)$ returns the probability density of a different class in the dataset, $r$, for feature $n$ at the same value. When calculating EC, class $c$ remains constant and is the class for which the instance has been predicted by the classifier. Where a feature is categorical, the function is calculated as follows is calculated using Eq.~\ref{eq:CED_cat}, as:

\begin{equation}\label{eq:CED_cat}
    f(x_n) = \frac{|z \in X_n^c \wedge X_n^c = x_n|}{|X^c|}
\end{equation}

\noindent where z refers to the total instances in the dataset, $X$, of class $c$, and feature $n$, where the category of the instance is the same as the category of instance $x$.

Given that difference features carry different levels of importance to the classifier, with some being more useful than others, the conflict matrix is weighted according to the Fishers Discriminant Ratio value for the features, normalised to highest ratio among the features. The total conflict score is calculated as the sum of the conflict across all features for each class comparison. The maximum conflict score across all class comparisons is returned.

\subsection{Proposed Uncertainty Estimation Framework}
\begin{figure}[ht]
    \centering
    \includegraphics[scale=0.29]{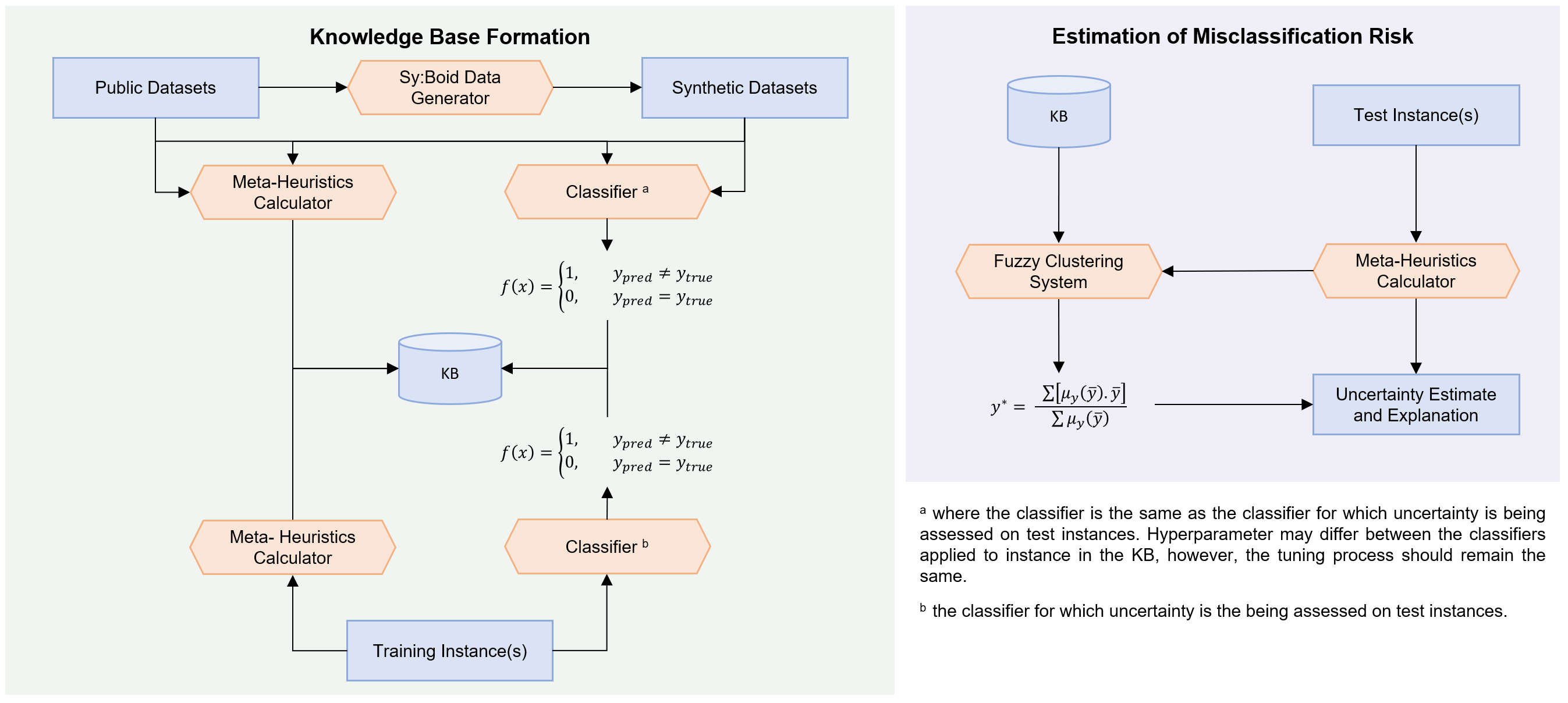}
    \caption{A graphical illustration of the framework for forming the knowledge base from which is used to train uncertainty estimation system.}
    \label{fig:Estimation_Framework}
\end{figure}

\noindent The proposed framework for estimating uncertainty is shown in Fig.~\ref{fig:Estimation_Framework}.

\subsubsection*{Knowledge Base Formation}
\noindent To train the fuzzy clustering system, a knowledge base is formed from the meta-data of real instances from the training set and other classification tasks. However, given that datasets from public repositories have drawn criticism for not being representative of the classification problems that may exist in the real world~\cite{macia2014towards} and that their homogeneity, from a complexity standpoint, can result in a biasing of algorithm development, the framework embeds a synthetic data generator, termed Sy:Boid~\cite{houston2022genetically}.

The process for generation is described fully in Houston and Cosma~\cite{houston2022genetically}, but briefly, the synthetic data generator takes inputs relating to the number of instances, number of features, number of classes, and the desired complexity, in this paper the F1 and N1 measures of  Ho
338 and Basu~\cite{ho2002complexity} are used. A series of random points are generated and labelled, and then, using a modified boid algorithm points are moved about the feature space and the complexity of the dataset is measured. The algorithm is embedded within a genetic algorithm which optimises the rules weightings to generate a dataset that closest meets the desired complexity. 

To generate synthetic datasets to complement the real data within the uncertainty system, Sy:Boid was tasked with generating datasets for all combinations of F1 and N1, where both F1 and N1 are values between 0 and 1, N1 is not larger than F1 as this rarely occurs in real-data~\cite{houston2022genetically}, and both F1 and N1 increase in increments of 0.2. The dataset generation was repeated for each dataset in table~\ref{tab:Dataset Description}, mimicking the number of instances, dimensionality and the number of classes. After all datasets were generated, 18020 instances were randomly selected, equaling the number of real instances.

The meta-heuristics for each instance in the training set, public datasets and synthetic dataset are calculated, using 5-fold cross-validation, where the meta-heuristics of the validation set are stored in the knowledge base. A single prediction is also provided for each instance by the classifier of interest, trained using a nested 5-fold cross-validation where the inner loop is used to determine the optimal hyper-parameters and the outer-loop is used to predict the class of the validation set. The outer loop of the cross-validation for evaluating the performance of the classifier is seeded to ensure the folds are identical to those used to calculate the meta-heuristics. Misclassification events are identified using Eq.~\ref{eq:misclassification_ID}, as:

\begin{equation}\label{eq:misclassification_ID}
    f(x) = \begin{cases}
        1 & y_{pred} \neq y_{true}\\
        0 & y_{pred}=y_{true}
    \end{cases}
\end{equation}

\noindent where $y_{pred}$ is the predicted class of an instance, and $y_{true}$ is the actual class of an instance. The misclassification events are then stored in the knowledge base with their respective meta-heuristics.

\subsubsection*{Uncertainty Estimation}
\noindent To examine the utility of the proposed measures in prospectively characterising the complexity of a patient, a weighted fuzzy c-means clustering approach is applied. Within the clustering system, the misclassification rate is calculated for each cluster using all instances contained within each respective cluster. Misclassification rates were calculated using Eq.~\ref{misclassification_eq}. 

\begin{equation}\label{misclassification_eq}
    Misclassification~Rate = 1-\frac{TP+TN}{{TP + FP + TN + FN} }
\end{equation}

\noindent where $TP$ refers to true positive prediction, $FN$ refers to a false negative prediction, $FP$ refers to false positive prediction and $TN$ refers to a true negative prediction.

When a new patient's data is input into the system their fuzzy membership is calculated, returning the membership of that patient to each cluster. To obtain a single value output, the output is defuzzified using the weighted average method, by:

\begin{equation}
    x^* = \frac{\sum[\mu_x(\bar{x})\times\bar{x}]}{\sum\mu_x(\bar{x}) }
\end{equation}

\noindent where $\mu_x$ is the multiplication of each weighting function of instance $x$ and $\bar{x}$ is the misclassification rate associated with each membership value.

To train and tune the clustering system, a 5-fold nested cross-validation approach is applied. The meta-heuristics proposed in subsection~\ref{subsec:meta-heuristics} were shown to vary in their strengths of association with misclassification performance (Fig.~\ref{fig:independent_associative_analysis}). Therefore, to improve the performance of the clustering system above that of one where each heuristic is weighted equally, Bayesian optimisation is applied within the inner-loop to determine the optimal weightings of the heuristics and number of clusters to include in the model, in order to maximise the product of the odds ratio, area under the receiver operating characteristic curve (AUROC) and area under the precision-recall curve (AUPRC), where the input is the defuzzified output and the target is the binary misclassification event. The outer-loop is used to evaluate the quality of the uncertainty estimates for identifying misclassification. The complete process for developing and training the predictive model and fuzzy clustering system is detailed in Algorithm~\ref{alg:External Eval Pseudo-code}. 

\begin{algorithm}[]
\small
\SetAlgoLined
\caption{Pseudo-code of the method for developing and training the predictive model and fuzzy clustering system.}
\label{alg:External Eval Pseudo-code}
\textbf{Inputs:} $X$ Dataset features, $Y$ Dataset Targets, $KB$ Knowledge Base\\
Divide $X$ and $Y$ into $K$ stratified folds\\ 
\For {$k_i$ in $K$ folds}{
Let $X_{k_i}$ and $Y_{k_i}$ be the test set features and targets, respectively\\

\# Generate classification model, model predictions and meta-heuristics for the training and testing set\\
\For {$k_j$ in $K-1$ folds}{
Let $X_{k_j}$ and $Y_{k_j}$ be the validation set features and targets, respectively\\
Train and tune model on remaining $K-2$ folds using Bayesian cross-validation, selecting parameters which maximise the balanced accuracy score.\\
Retrain optimal model on all $K-2$ folds.\\
Let $y_{pred}$ be the predicted class of the trained model applied to $X_{k_j}$\\
Let $M_{k_j} = y_{pred} != Y_{k_j}$\\
Let $C_{k_j}$ be the calculate meta-heuristics for $X_{k_j}$ with the remaining $K-2$ folds acting as the training set.
}
Retrain optimal model on all $K-1$ folds
Let $y_{pred}$ be the predicted class of the trained model applied to $X_{k_i}$\\
Let $M_{k_i} = y_{pred} != Y_{k_i}$\\
Let $C_{k_i}$ be the calculate meta-heuristics for $X_{k_i}$ with the remaining $K-1$ folds acting as the training set.

\# Generate clustering system\\
\For {$k_j$ in $K-1$ folds}{
Let $C_{k_j}$ and $M_{k_j}$ be the validation complexity heuristics and misclassification targets, respectively\\
Train and tune clustering system on remaining $K-2$ folds and $KB$ using Bayesian cross-validation selecting the parameter weights and number of clusters which maximise the fitness function on the validation set\\
}
Let $C_{k_i}$ = $C_{k_i} \times$ optimal weights\\
Retrain optimal clustering system on all $K-1$ folds and $KB$\\
Let $U_{k_i}$ be the estimated uncertainty of the trained clustering systems defuzzified output for all instances in $C_{k_i}$.\\
}
\# Evaluate Performance\\
Evaluate performance over all $K$ folds using $U$ and $M$
\end{algorithm}
\normalsize

\section{Experimental Methods}\label{sec:Experimental_Methods}
\noindent This section describes the experimental methods for determining the optimal design of the uncertainty estimation system and evaluating its performance on unseen data.

\subsection{Dataset Description}
\noindent Twenty-seven publicly available datasets were used in experiments to evaluate the statistical relationships between the proposed meta-features and misclassification events, and to determine the optimal design of the knowledge base. A description of the datasets and their source is provided in Table~\ref{tab:Dataset Description}. The datasets vary in their number of instances and dimensionality (i.e. the number of features). The selected datasets include continuous, ordinal and categorical features, to ensure the proposed meta-heuristics and developed system for estimating uncertainty are evaluated across a range of datasets.

{\setstretch{1.0}
\begin{table}[ht]
    \footnotesize
    \centering
    \centerline{
    \begin{tabular}{|c|c|ccc|ccc|}
        \hline
        \multirow{2}{*}{\textbf{Dataset}} &
          \multirow{2}{*}{\textbf{Source}} &
          \multicolumn{3}{c|}{\textbf{Size}} &
          \multicolumn{3}{c|}{\textbf{Feature Types}} \\ \cline{3-8} 
         &
           &
          \multicolumn{1}{c|}{\textbf{Instances}} &
          \multicolumn{1}{c|}{\textbf{Dims.}} &
          \textbf{Classes} &
          \multicolumn{1}{c|}{\textbf{Nominal}} &
          \multicolumn{1}{c|}{\textbf{Ordinal}} &
          \textbf{Continuous}  \\ \hline
          Acute Inflammations (Inflammation) &
          \cite{czerniak2003application} &
          \multicolumn{1}{c|}{120} &
          \multicolumn{1}{c|}{6} &
          2 &
          \multicolumn{1}{c|}{Y} &
          \multicolumn{1}{c|}{N} &
          Y \\ \hline
        Acute Inflammations (Nephritis) &
          \cite{czerniak2003application} &
          \multicolumn{1}{c|}{120} &
          \multicolumn{1}{c|}{6} &
          2 &
          \multicolumn{1}{c|}{Y} &
          \multicolumn{1}{c|}{N} &
          Y \\ \hline
        Breast Cancer-C &
          \cite{patricio2018using} &
          \multicolumn{1}{c|}{116} &
          \multicolumn{1}{c|}{9} &
          2 &
          \multicolumn{1}{c|}{N} &
          \multicolumn{1}{c|}{N} &
          Y \\ \hline
        Breast Cancer-W &
          \cite{Dua:2019} &
          \multicolumn{1}{c|}{699} &
          \multicolumn{1}{c|}{9} &
          2 &
          \multicolumn{1}{c|}{N} &
          \multicolumn{1}{c|}{N} &
          Y  \\ \hline
        Breast Tissue &
          \cite{Dua:2019} &
          \multicolumn{1}{c|}{106} &
          \multicolumn{1}{c|}{9} &
          6 &
          \multicolumn{1}{c|}{N} &
          \multicolumn{1}{c|}{N} &
          Y  \\ \hline
        CECS &
          \cite{houston2021predicting} &
          \multicolumn{1}{c|}{126} &
          \multicolumn{1}{c|}{23} &
          2 &
          \multicolumn{1}{c|}{Y} &
          \multicolumn{1}{c|}{Y} &
          Y  \\ \hline
        C-Section &
          \cite{Dua:2019} &
          \multicolumn{1}{c|}{80} &
          \multicolumn{1}{c|}{5} &
          2 &
          \multicolumn{1}{c|}{Y} &
          \multicolumn{1}{c|}{N} &
          Y  \\ \hline
        Chronic Kidney Disease &
          \cite{Dua:2019} &
          \multicolumn{1}{c|}{351} &
          \multicolumn{1}{c|}{24} &
          2 &
          \multicolumn{1}{c|}{Y} &
          \multicolumn{1}{c|}{Y} &
          Y  \\ \hline
        Diabetic Retinopathy &
          \cite{antal2014ensemble} &
          \multicolumn{1}{c|}{1151} &
          \multicolumn{1}{c|}{18} &
          2 &
          \multicolumn{1}{c|}{Y} &
          \multicolumn{1}{c|}{N} &
          Y  \\ \hline
        Early Stage Diabetes &
          \cite{islam2020likelihood} &
          \multicolumn{1}{c|}{520} &
          \multicolumn{1}{c|}{16} &
          2 &
          \multicolumn{1}{c|}{Y} &
          \multicolumn{1}{c|}{N} &
          Y  \\ \hline
        Echocardiogram &
          \cite{Dua:2019} &
          \multicolumn{1}{c|}{126} &
          \multicolumn{1}{c|}{8} &
          2 &
          \multicolumn{1}{c|}{Y} &
          \multicolumn{1}{c|}{N} &
          Y  \\ \hline
        Fertility &
          \cite{gil2012predicting} &
          \multicolumn{1}{c|}{100} &
          \multicolumn{1}{c|}{9} &
          2 &
          \multicolumn{1}{c|}{Y} &
          \multicolumn{1}{c|}{Y} &
          Y  \\ \hline
        HCV & 
          \cite{Dua:2019} &
          \multicolumn{1}{c|}{615} &
          \multicolumn{1}{c|}{12} &
          2 &
          \multicolumn{1}{c|}{Y} &
          \multicolumn{1}{c|}{N} &
          Y  \\ \hline
        Heart Disease-C & 
          \cite{Dua:2019} &
          \multicolumn{1}{c|}{303} &
          \multicolumn{1}{c|}{13} &
          2 &
          \multicolumn{1}{c|}{Y} &
          \multicolumn{1}{c|}{Y} &
          Y  \\ \hline
        Heart Failure &
          \cite{chicco2020machine} &
          \multicolumn{1}{c|}{299} &
          \multicolumn{1}{c|}{12} &
          2 &
          \multicolumn{1}{c|}{Y} &
          \multicolumn{1}{c|}{N} &
          Y  \\ \hline
        Hepatitis &
          \cite{Dua:2019} &
          \multicolumn{1}{c|}{154} &
          \multicolumn{1}{c|}{19} &
          2 &
          \multicolumn{1}{c|}{Y} &
          \multicolumn{1}{c|}{N} &
          Y  \\ \hline
        Hypothyroid &
          \cite{Dua:2019} &
          \multicolumn{1}{c|}{3163} &
          \multicolumn{1}{c|}{19} &
          2 &
          \multicolumn{1}{c|}{Y} &
          \multicolumn{1}{c|}{N} &
          Y  \\ \hline
        Indian Liver Disease &
          \cite{Dua:2019} &
          \multicolumn{1}{c|}{583} &
          \multicolumn{1}{c|}{10} &
          2 &
          \multicolumn{1}{c|}{Y} &
          \multicolumn{1}{c|}{N} &
          Y \\ \hline
        LSVT &
          \cite{tsanas2013objective} &
          \multicolumn{1}{c|}{127} &
          \multicolumn{1}{c|}{312} &
          2 &
          \multicolumn{1}{c|}{Y} &
          \multicolumn{1}{c|}{N} &
          Y  \\ \hline
        Lymphography &
          \cite{Dua:2019} &
          \multicolumn{1}{c|}{142} &
          \multicolumn{1}{c|}{36} &
          2 &
          \multicolumn{1}{c|}{Y} &
          \multicolumn{1}{c|}{N} &
          Y  \\ \hline
        Maternal Health Risk &
          \cite{Dua:2019} &
          \multicolumn{1}{c|}{1014} &
          \multicolumn{1}{c|}{6} &
          3 &
          \multicolumn{1}{c|}{N} &
          \multicolumn{1}{c|}{N} &
          Y  \\ \hline
        Myocardial Infarct (Heart Failure) &
          \cite{golovenkin2020trajectories} &
          \multicolumn{1}{c|}{1700} &
          \multicolumn{1}{c|}{111} &
          2 &
          \multicolumn{1}{c|}{Y} &
          \multicolumn{1}{c|}{Y} &
          Y  \\ \hline
        Myocardial Infarct (Death) &
          \cite{golovenkin2020trajectories} &
          \multicolumn{1}{c|}{1700} &
          \multicolumn{1}{c|}{111} &
          2 &
          \multicolumn{1}{c|}{Y} &
          \multicolumn{1}{c|}{Y} &
          Y  \\ \hline
        Myocardial Infarct (Infarction) &
          \cite{golovenkin2020trajectories} &
          \multicolumn{1}{c|}{1700} &
          \multicolumn{1}{c|}{111} &
          2 &
          \multicolumn{1}{c|}{Y} &
          \multicolumn{1}{c|}{Y} &
          Y  \\ \hline
        NKI Breast &
          \cite{van2002gene} &
          \multicolumn{1}{c|}{272} &
          \multicolumn{1}{c|}{12} &
          2 &
          \multicolumn{1}{c|}{Y} &
          \multicolumn{1}{c|}{Y} &
          Y  \\ \hline
        Parkinsons Speech &
          \cite{sakar2019comparative} &
          \multicolumn{1}{c|}{756} &
          \multicolumn{1}{c|}{753} &
          2 &
          \multicolumn{1}{c|}{Y} &
          \multicolumn{1}{c|}{N} &
          Y  \\ \hline
        Pancreatic Cancer &
          \cite{debernardi2020combination} &
          \multicolumn{1}{c|}{590} &
          \multicolumn{1}{c|}{8} &
          3 &
          \multicolumn{1}{c|}{Y} &
          \multicolumn{1}{c|}{N} &
          Y  \\ \hline
        Parkinsons &
          \cite{little2008suitability} &
          \multicolumn{1}{c|}{195} &
          \multicolumn{1}{c|}{22} &
          2 &
          \multicolumn{1}{c|}{N} &
          \multicolumn{1}{c|}{N} &
          Y  \\ \hline
        Thoracic Surgery &
          \cite{zikeba2014boosted} &
          \multicolumn{1}{c|}{470} &
          \multicolumn{1}{c|}{17} &
          2 &
          \multicolumn{1}{c|}{Y} &
          \multicolumn{1}{c|}{Y} &
          Y  \\ \hline
        Vertebral Column (2-Class) &
          \cite{Dua:2019} &
          \multicolumn{1}{c|}{311} &
          \multicolumn{1}{c|}{6} &
          2 &
          \multicolumn{1}{c|}{N} &
          \multicolumn{1}{c|}{N} &
          Y  \\ \hline
        Vertebral Column (3-Class) &
          \cite{Dua:2019} &
          \multicolumn{1}{c|}{311} &
          \multicolumn{1}{c|}{6} &
          3 &
          \multicolumn{1}{c|}{N} &
          \multicolumn{1}{c|}{N} &
          Y  \\ \hline
    \end{tabular}}
    \caption{Description of the datasets included in this paper.}
    \label{tab:Dataset Description}
    \normalsize
\end{table}}

\subsubsection{Statistical Evaluation of the Proposed Meta-Heuristics}
\noindent The intention of each meta-heuristic is to characterise the complexity of an instance, with the sources of complexity being independent from each other. Therefore, to assess the independence of each meta-heuristic, a correlation analysis was performed using Spearman's Rank Correlation. To assess the association of each meta-heuristic with misclassifications, independent of each other, univariate binary logistic regressions were applied for each model, with misclassifications (binary yes/no) acting as the dependent variable and each meta-heuristic acting as the independent variable. Prior to applying the logistic regression, each meta-heuristic underwent z-score transformation to aid in the of interpretation of the odds-ratios.

\subsection{Statistical Evaluation of the Knowledge Base Construction}
\noindent To generate the optimal knowledge base, three things must be known: 1) how many instances should be included, 2) how diverse from the current instance should the selected instances be, 3) should the knowledge base be comprised of real instances, synthetic instances or a combination of both. To identify the optimal parameters, for each patient, 48 knowledge bases were generated comprised of $m$ instances randomly selected from the nearest $q$\% of instances to the current instance, where $m$ = 100, 500, 1000 or 2000, and $q$ = 10\%, 25\%, 50\% and 100\%, with these instances selected from a meta-database comprised of all real instances, all synthetic instances and a combination of real and fake instances. The fuzzy clustering system was trained, according to Algorithm~\ref{alg:External Eval Pseudo-code}, on each of the knowledge bases each time generating 10 defuzzified outputs for each instance (one for each classifier). For each model, an odds ratio, AUROC and AUPRC were calculated for the outputs from the fuzzy clustering system, trained on each of the 48 knowledge bases.

To assess what parameters resulted in the defuzzified outputs being best suited for identifying misclassification, ANOVAs were applied with the calculated odds-ratio, AUROC, AUPRC acting as the dependent variables and the number of instances, diversity of the selected instances and realness of the knowledge base acting as the independent variables. Additionally, the interaction effect was calculated between the number of instances and the diversity of the selected instances. Where significant main effects were found, post-hoc analyses were performed for all two-variable combinations, apply Bonferroni's correction to account for multiple comparisons. For all statistical tests, the null-hypothesis was rejected when $\alpha < 0.05$.

\subsection{External Validation of the Proposed Methods}
\noindent To ensure the performance of the proposed methods are evaluated on unseen data, not used to identify the optimal knowledge base, the dataset described in Hou et al.~\cite{hou2020predicting}, comprised of 4559 patients with sepsis-3 from the MIMIC-III database was replicated. The full process for extracting and pre-processing the patient data is detailed in Hou et al.~\cite{hou2020predicting}. The uncertainty system was formed and evaluated according to Algorithm~\ref{alg:External Eval Pseudo-code}, with performance measured in terms of the odds ratio between estimated uncertainty and actual misclassification events, AUROC and AUPRC.

Given that the ability to identify misclassifications without class knowledge is limited to instances where there is at least some information that allows them to be classified correctly. The evaluation of the uncertainty estimations is repeated twice. First including all patients, then again, removing instances where their presentation and class target means they should be misclassified. To determine instances which should be misclassified (ISMs), the methods of Smith and Martinez~\cite{smith2011improving} are applied. First, each instance is characterised in terms of DS, Disjunct Class Percentage (DCP) and Class-Likelihood Difference (CLD), where Disjunct size is calculated using Eq.~\ref{eq:DS}, and DCP and CLP are calculated using Eqs.~\ref{eq:DCP} and~\ref{eq:CLD}, as:
\begin{equation}\label{eq:DCP}
    DCP(x) = \frac{|\{z:z \in disjunct(x) \wedge t(z) = t(x)\}|}{|disjunct(x)|}
\end{equation}
\begin{equation}\label{eq:CLD}
    CLD(x) = CL(x, t(x)) - \underset{y \in Y - t(x)}{argmaxCL(x,y)}
\end{equation}

\noindent where the function $disjunct(x)$ returns the disjunct that covers instance $x$, the function $t(x)$ returns the class label of instance $x$, $z$ refers to instances contained in the returned disjunct, $Y$ is a set containing each unique class label in the dataset and the function $CL(x)$ returns the probability of an instance belonging to a certain class, and is derived using Eq.~(\ref{CL}):

\begin{equation}\label{CL}
    CL(x) = CL(x,t(x)) = \prod_{n}^{N}P(x_n|t(x))
\end{equation}

\noindent where $x_n$ is the $n$th feature of instance $x$. Thereafter, should an instance meet the following condition $CLD(x, t(x)) < 0~$and$~((DS(x) == 0~$and$~DCP(x) < 0.5)~$or$ DN(x) > 0.8)$, the are considered an ISM.

Due to the increased training time required to calculate the meta-heuristics and train the fuzzy-clustering system, the performance of the estimated uncertainty output by the fuzzy clustering system is compared against the absolute difference of the classification probability for the predicted class from 0.5. For all models, except SVM, classification probability is calculated as the conditional probability of the predicted class. For SVM classification probability is calculated using Platt scaling~\cite{platt1999probabilistic}.

\section{Experimental Results}\label{sec:Experimental_Results}

\subsection{Statistical Analysis of the Proposed Meta-Heuristics}

\subsubsection*{Investigation of Co-Linearity between Meta-Heuristics}
\begin{figure}[ht]
    \centering
    \centerline{\includegraphics[scale=0.5]{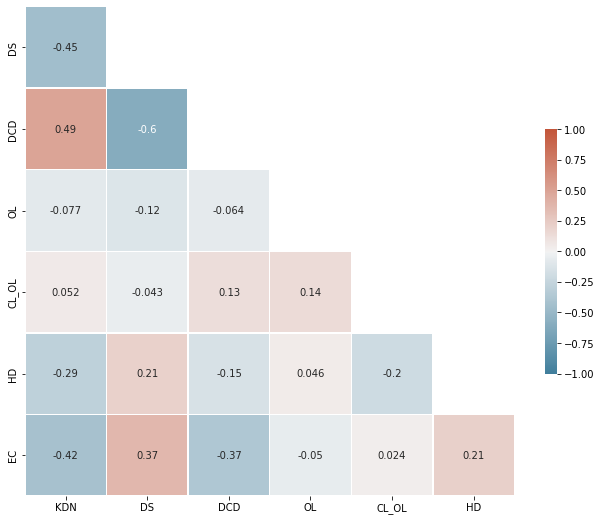}}
    \caption{Spearman's rank correlation matrix showing the relationship between each meta-heuristic. A negative value, highlighted in blue, represented a negative correlation and a positive value, highlighted in red, represents a positive correlation.}
    \label{fig:correlation matrix}
\end{figure}

\noindent Results of the correlation analysis are presented in Fig.~\ref{fig:correlation matrix}. Results showed minimal co-linearity between meta-heuristics, meaning the source of uncertainty that each meta-heuristic measures are independent of one and other. The only pair of meta-heuristics which demonstrated co-linearity was DS and DCD ($\rho =$ -0.60). However, due to both DS and DCD being disjunct-based measures, it is not unsurprising that a degree of overlap exists within the domains of uncertainty they capture.

\subsubsection*{Investigation of the Independent Associations between Meta-Heuristics and Misclassification Events}
\begin{figure}[ht]
    \centering
    \centerline{\includegraphics[scale=0.65]{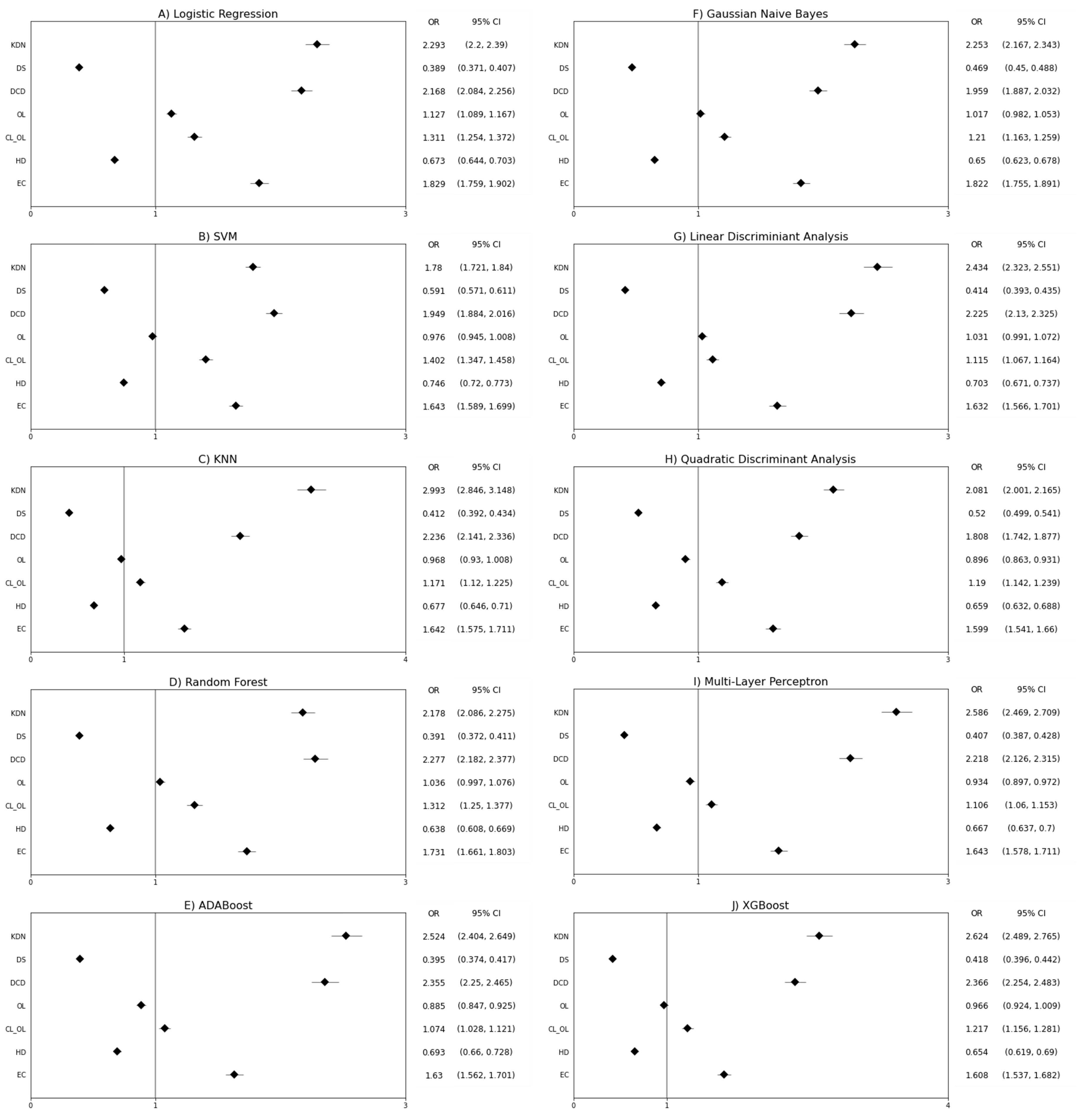}}
    \caption{Odds ratios and respective 95\% confidence intervals demonstrating the independent association of each meta-heuristic a misclassification. Diamonds indicate the odds ratio, with an odds ratio greater than one reflecting a positive association with a misclassification.}
    \label{fig:independent_associative_analysis}
\end{figure}

\noindent Results of the association-based analyses are presented in Fig.~\ref{fig:independent_associative_analysis}. Results indicate significant associations across all classifiers for almost all heuristics. A summary of the figure is provided below:

\begin{itemize}
    \item \textit{KDN: }An increase in KDN was observed to significantly increase the odds of a misclassification occurring across all classifiers (p $<$ 0.001), with odds ratios ranging from 1.780 (95\% CI = 1.721 - 1.84) for SVM to 2.993 (95\% CI = 2.846 - 1.84) for KNN, meaning that the more diverse the nearby instances are, in terms of their class labels, the greater chance of a misclassification occurring.
    \item \textit{DS: }An increase in DS was shown to significantly decrease the likelihood of a misclassification occurring across all classifiers (p $<$ 0.001), with odds ratios ranging from 0.389 (95\% CI = 0.371 - 0.407) for logistic regression, to 0.591 (95\% CI = 0.371 - 0.407) for SVM, meaning that the more complex the decision boundary for a given instance, relative to that of other instances, the greater the chance of misclassification occurring. 
    \item \textit{DCD: }An increase in DCD was observed to significantly increase the odds of a misclassification occurring across all classifiers (p $<$ 0.001), with odds ratios ranging from 1.808 (95\% CI = 1.742 - 1.877) for QDA to 2.366 (95\% CI = 2.254 - 2.483) for XGBoost, meaning that the more diverse the nearby instances are, in terms of their class labels, based on a subset of the available features, the greater chance of a misclassification occurring. 
    \item \textit{OL: }An increase in OL affected the likelihood of misclassification occurring differently for each classifier. For logistic regression, an increase in OL significantly increased the odds of a misclassification occurring (OR = 1.127, 95\% CI = 1.089 - 1.167), that is to say, the more outlying an instance from the instances in the training set, the greater chance of a misclassification. Whereas, for Adaboost, QDA and MLP, the opposite is true, with an increase in OL significantly decreasing the chance of misclassification occurring. Increases in OL was shown to not significantly affect the likelihood of a misclassification occurring for SVM, KNN, Random Forest, LDA or XGBoost (p $>$ 0.05). 
    \item \textit{CL-OL: }An increase in CL-OL was found to significantly increase the likelihood of a misclassification occurring across all classifiers (p $<$ 0.05), with odds ratios ranging from 1.017 (95\% 1.028 - 1.121) for Adaboost to 1.402 (95\% CI 1.347 - 1.458) for SVM, meaning that the more equal an instance's outlierness is to all available classes, the greater the chance of a misclassification occurring. 
    \item \textit{HD: }An increase in HD was shown to significantly decrease the likelihood of a misclassification occurring across all classifiers (p $<$ 0.001), with odds ratios ranging from 0.638 (95\% CI = 0.606 - 0.669) for Random Forest, to 0.746 (95\% CI = 0.720 - 0.773) for SVM, meaning that the closer to the hyperplane an instance falls, the greater the chance of a misclassification occurring. 
    \item \textit{EC: }An increase in EC was observed to significantly increase the odds of a misclassification occurring across all classifiers (p $<$ 0.001), with odds ratios ranging from 1.599 (95\% CI = 1.541 - 1.660) for QDA to 1.829 (95\% CI 1.759 - 1.902) for logistic regression, meaning that the degree to which the features of an instance point to two potential classes, the greater the chance of a misclassification occurring.
\end{itemize}

\subsubsection*{Summary}
\noindent Results of the statistical analysis demonstrate that each of the proposed meta-heuristics characterises instances in terms of different aspects of uncertainty, independent of each other. However, as demonstrated by the association-based analysis, some meta-heuristics are more strongly associated with misclassifications than others and therefore, should not be weighted equally when characterising the overall uncertainty of a classification.

\subsection{Identification of the Optimal Knowledge Base}

\subsubsection*{Effect of Size}
\noindent Results of the between-group analysis found a significant difference in the odd ratios obtained from knowledge bases of different sizes (p $<$ 0.001), with the odds ratios obtained using knowledge bases comprised of only 100 instances (2.29 $\pm$ 0.36) being significantly lower (p = 0.001) than those obtained by knowledge bases containing 500 (2.70 $\pm$ 0.41), 1000 (2.85 $\pm$ 0.47) and 2000 instances (2.93 $\pm$ 0.50). Odds ratios obtained using a knowledge base comprised of 500 instances were significantly lower than those obtained by knowledge bases comprised of 1000 and 2000 instances (p = 0.045 and 0.001, respectively). No significant difference was observed in the odds ratios obtained from knowledge bases comprised of 1000 and 2000 instances (p = 0.502). This observation was mirrored in the results of the analysis of both the AUROC and AUPRC (p $<$ 0.001 and p = 0.027, respectively), with AUROCs being significantly lower when only 100 instances were used (0.73 $\pm$ 0.03) compared to when 500 (0.75 $\pm$ 0.03), 1000 (0.75 $\pm$ 0.03) and 2000 (0.75 $\pm$ 0.03) were used (p = 0.001) and AUPRC being significantly worse when the knowledge base contained only 100 instances compared to 2000 (0.33 $\pm$ 0.06 vs. 0.35 $\pm$ 0.06, p = 0.033).

\subsubsection*{Effect of Diversity}
\noindent In terms of the diversity of the knowledge base, the between-group analysis found significant differences between the four groups for the analysis of odd ratios (p $<$ 0.001). However, post-hoc analysis reveals the only significant difference (p = 0.001) in odds ratio was that knowledge bases formed using the nearest 50\% of instances to the current instance (2.43 $\pm$ 0.37), were significantly worse than those formed by the nearest 10\% (2.82 $\pm$ 0.40), 25\% (2.77 $\pm$ 0.40) and 100\% (2.76 $\pm$ 0.68). Similarly, there was a significant difference in the AUROCs achieved by different thresholds (p $<$ 0.001). The post-hoc analysis revealed that the smaller thresholds of 10\% and 25\% resulted in significantly higher (p $<$ 0.01) AUROCs (0.75 $\pm$ 0.03) than the larger thresholds of 50\% and 100\% (0.74 $\pm$ 0.03 and 0.73 $\pm$ 0.04). No significant effect was observed in terms of AUPRC (p = 0.095)

\subsubsection*{Interaction Between Size and Diversity}
\begin{figure}[ht]
    \centering
    \centerline{\includegraphics[scale=0.8]{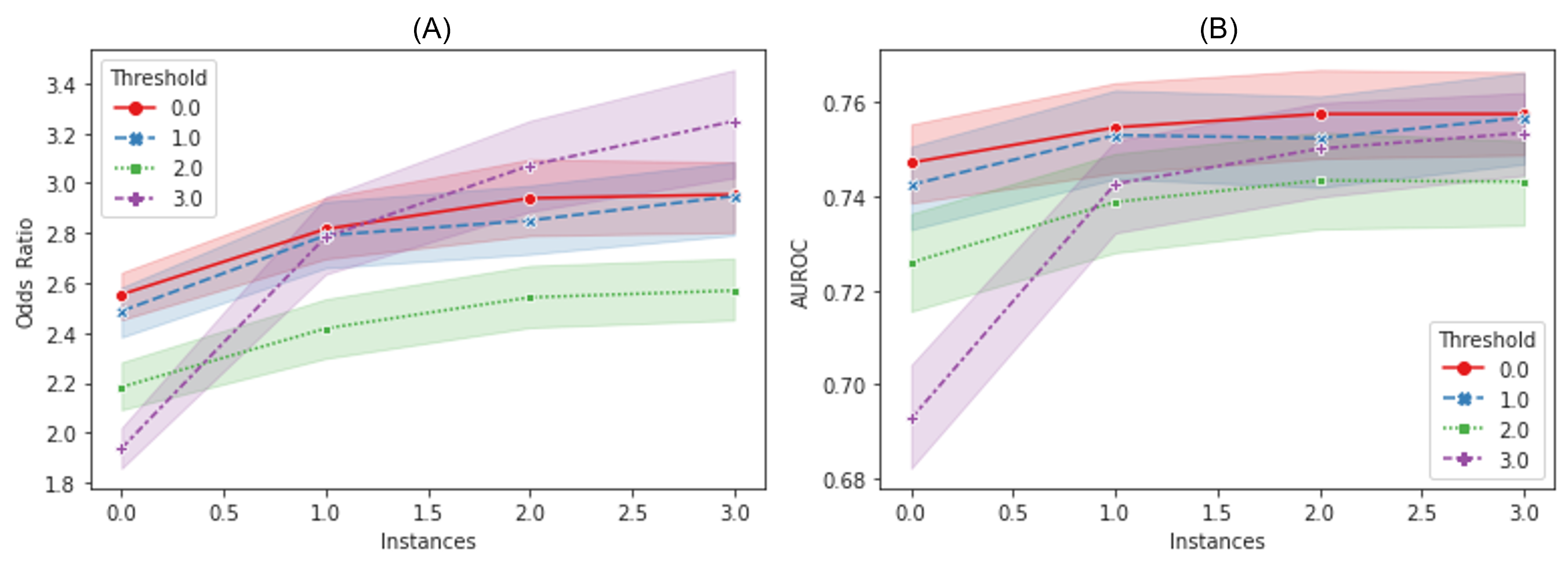}}
    \caption{The interaction for (A) odds-ratio and (B) AUROC, between the number of instances included in the knowledge base and the diversity of the knowledge base, where each line reflects the different diversity groups.}
    \label{fig:Between-Group Interaction}
\end{figure}

\noindent An interaction effect was observed between the number of instances included in the knowledge base and the diversity of the knowledge base for both odd-ratios and AUROCs (p $<$ 0.001). As shown in Fig.~\ref{fig:Between-Group Interaction}, when the number of instances is small, the performance of highly diverse knowledge bases suffers, however as the number of instances increases, performance improves. In the case of knowledge bases comprised of less diverse instances, performance plateaus as more instances are introduced, which is likely the result of a saturated meta-feature space. No interaction was observed in terms of AUPRC (p = 0.670).

\subsubsection*{Effect of Realness}
\noindent Concerning the realness of the knowledge base, between-group differences were observed in terms of odds ratios (p = 0.003). Post-hoc analyses revealed that knowledge bases comprised solely of fake instances led to significantly higher (p = 0.036) odds ratios (2.750 $\pm$ 0.50) than those of solely real instances (2.611 $\pm$ 0.51). No significant differences were observed between the knowledge bases comprised of both real and fake instances (2.719 $\pm$ 0.49) and the other two groups (p $>$ 0.05). The realness of the knowledge base did not affect the performance in terms of either AUROC or AUPRC (p = 0.210 and 0.242, respectively).

\subsubsection*{Summary}
\noindent Due to the interaction effect between the threshold and the number of instances, all experiments will be carried out using a threshold of 100\% and selecting 2000 instances to train the fuzzy clustering system. Additionally, due to the desire to retain some reality within the knowledge base, and because of negligible performances differences between the knowledge bases comprised of fake and a combination of real and fake instances, the knowledge base will be formed of both real and synthetic instances.

\subsection{External Validation of the Proposed Uncertainty Estimation System}\label{subsec:External_Eval}
\noindent Table~\ref{tab:performance} shows the performance of the proposed method, with and without the inclusion of ISMs, compared against the use of classifier probabilities. Results show that in terms of odds ratios, the proposed method proved superior for 8 of the 10 models when ISMs were included and 5 of the 10 models when classifiers were excluded, demonstrating that the estimated uncertainty is equally associated with the classifier probabilities for instances which should not be misclassified. In terms of AUROC, both with and without ISMs included the proposed method was superior in 9 out of 10 models, with the only model for which the classifier probabilities proved superior being SVM, demonstrating that the uncertainty estimates achieved through the proposed method achieve greater discriminative performance between positive and negative examples, across the majority of models than classifier uncertainty. Lastly, in terms of AUPRC, the proposed method proved superior in 8 out of 10 models when ISMs were included and 6 out of 10 models when ISMs were removed, demonstrating the proposed methods are better at identifying misclassifications than classifier probability across more models both with and without the presence of ISMs. Generally, the proposed methods proved more robust in providing estimations of uncertainty in the presence of ISMs, however, the performance of both methods was improved by their removal.

\begin{landscape}
\begin{table}[]
\small
\centering
\begin{tabular}{|ccccccc|}
\hline
\multicolumn{1}{|l|}{} &
  \multicolumn{3}{c|}{Proposed Method} &
  \multicolumn{3}{c|}{Classifier   Probabilites} \\ \hline
\multicolumn{1}{|c|}{Model} &
  \multicolumn{1}{c|}{Odds Ratio (95\% CI)} &
  \multicolumn{1}{c|}{AUROC} &
  \multicolumn{1}{c|}{\begin{tabular}[c]{@{}c@{}}AUPRC \\ (Improvement from Random)\end{tabular}} &
  \multicolumn{1}{c|}{Odds Ratio (95\% CI)} &
  \multicolumn{1}{c|}{AUROC} &
  \begin{tabular}[c]{@{}c@{}}AUPRC \\ (Improvement from Random)\end{tabular} \\ \hline
\multicolumn{7}{|c|}{Including ISMs} \\ \hline
\multicolumn{1}{|c|}{LR} &
  \multicolumn{1}{c|}{\textbf{3.263 (2.993 - 3.558)}} &
  \multicolumn{1}{c|}{\textbf{0.79}} &
  \multicolumn{1}{c|}{\textbf{0.52 (0.27)}} &
  \multicolumn{1}{c|}{2.423 (2.231 - 2.633)} &
  \multicolumn{1}{c|}{0.72} &
  0.42 (0.18) \\ \hline
\multicolumn{1}{|c|}{SVM} &
  \multicolumn{1}{c|}{1.459 (1.372 - 1.552)} &
  \multicolumn{1}{c|}{0.61} &
  \multicolumn{1}{c|}{0.45 (0.08)} &
  \multicolumn{1}{c|}{\textbf{2.279 (2.117 - 2.454)}} &
  \multicolumn{1}{c|}{\textbf{0.71}} &
  \textbf{0.56 (0.19)} \\ \hline
\multicolumn{1}{|c|}{KNN} &
  \multicolumn{1}{c|}{\textbf{2.814 (2.588 - 3.06)}} &
  \multicolumn{1}{c|}{\textbf{0.76}} &
  \multicolumn{1}{c|}{\textbf{0.5 (0.23)}} &
  \multicolumn{1}{c|}{2.1 (1.955 - 2.255)} &
  \multicolumn{1}{c|}{0.70} &
  0.43 (0.16) \\ \hline
\multicolumn{1}{|c|}{RF} &
  \multicolumn{1}{c|}{2.743 (2.521 - 2.985)} &
  \multicolumn{1}{c|}{\textbf{0.76}} &
  \multicolumn{1}{c|}{\textbf{0.42 (0.22)}} &
  \multicolumn{1}{c|}{\textbf{3.08 (2.775 - 3.419)}} &
  \multicolumn{1}{c|}{0.75} &
  0.4 (0.2) \\ \hline
\multicolumn{1}{|c|}{ADA} &
  \multicolumn{1}{c|}{\textbf{2.819 (2.594 - 3.062)}} &
  \multicolumn{1}{c|}{\textbf{0.75}} &
  \multicolumn{1}{c|}{\textbf{0.47 (0.2)}} &
  \multicolumn{1}{c|}{2.438 (2.241 - 2.651)} &
  \multicolumn{1}{c|}{0.72} &
  0.44 (0.17) \\ \hline
\multicolumn{1}{|c|}{GNB} &
  \multicolumn{1}{c|}{\textbf{2.24 (2.07 - 2.424)}} &
  \multicolumn{1}{c|}{\textbf{0.72}} &
  \multicolumn{1}{c|}{\textbf{0.37 (0.16)}} &
  \multicolumn{1}{c|}{1.468 (1.372 - 1.571)} &
  \multicolumn{1}{c|}{0.59} &
  0.31 (0.11) \\ \hline
\multicolumn{1}{|c|}{LDA} &
  \multicolumn{1}{c|}{\textbf{3.139 (2.886 - 3.415)}} &
  \multicolumn{1}{c|}{\textbf{0.79}} &
  \multicolumn{1}{c|}{\textbf{0.51 (0.27)}} &
  \multicolumn{1}{c|}{2.45 (2.258 - 2.659)} &
  \multicolumn{1}{c|}{0.73} &
  0.43 (0.19) \\ \hline
\multicolumn{1}{|c|}{QDA} &
  \multicolumn{1}{c|}{\textbf{2.643 (2.442 - 2.86)}} &
  \multicolumn{1}{c|}{\textbf{0.76}} &
  \multicolumn{1}{c|}{\textbf{0.44 (0.22)}} &
  \multicolumn{1}{c|}{1.974 (1.833 - 2.125)} &
  \multicolumn{1}{c|}{0.69} &
  0.37 (0.15) \\ \hline
\multicolumn{1}{|c|}{MLP} &
  \multicolumn{1}{c|}{\textbf{2.568 (2.37 - 2.782)}} &
  \multicolumn{1}{c|}{\textbf{0.75}} &
  \multicolumn{1}{c|}{\textbf{0.42 (0.19)}} &
  \multicolumn{1}{c|}{2.543 (2.338 - 2.766)} &
  \multicolumn{1}{c|}{0.74} &
  0.41 (0.18) \\ \hline
\multicolumn{1}{|c|}{XGB} &
  \multicolumn{1}{c|}{2.409 (2.223 - 2.61)} &
  \multicolumn{1}{c|}{\textbf{0.74}} &
  \multicolumn{1}{c|}{0.36 (0.18)} &
  \multicolumn{1}{c|}{\textbf{2.495 (2.285 - 2.726)}} &
  \multicolumn{1}{c|}{\textbf{0.74}} &
  \textbf{0.37 (0.19)} \\ \hline
\multicolumn{7}{|c|}{Excluding ISMs} \\ \hline
\multicolumn{1}{|c|}{LR} &
  \multicolumn{1}{c|}{\textbf{4.711 (4.157 - 5.339)}} &
  \multicolumn{1}{c|}{\textbf{0.85}} &
  \multicolumn{1}{c|}{\textbf{0.49 (0.33)}} &
  \multicolumn{1}{c|}{3.991 (3.503 - 4.547)} &
  \multicolumn{1}{c|}{0.80} &
  0.39 (0.23) \\ \hline
\multicolumn{1}{|c|}{SVM} &
  \multicolumn{1}{c|}{1.434 (1.34 - 1.535)} &
  \multicolumn{1}{c|}{0.61} &
  \multicolumn{1}{c|}{0.42 (0.08)} &
  \multicolumn{1}{c|}{\textbf{2.784 (2.548 - 3.041)}} &
  \multicolumn{1}{c|}{\textbf{0.75}} &
  \textbf{0.58 (0.24)} \\ \hline
\multicolumn{1}{|c|}{KNN} &
  \multicolumn{1}{c|}{\textbf{3.537 (3.176 - 3.94)}} &
  \multicolumn{1}{c|}{\textbf{0.80}} &
  \multicolumn{1}{c|}{\textbf{0.46 (0.25)}} &
  \multicolumn{1}{c|}{2.705 (2.469 - 2.963)} &
  \multicolumn{1}{c|}{0.76} &
  0.41 (0.2) \\ \hline
\multicolumn{1}{|c|}{RF} &
  \multicolumn{1}{c|}{3.667 (3.273 - 4.108)} &
  \multicolumn{1}{c|}{\textbf{0.83}} &
  \multicolumn{1}{c|}{\textbf{0.4 (0.26)}} &
  \multicolumn{1}{c|}{\textbf{4.724 (4.051 - 5.508)}} &
  \multicolumn{1}{c|}{0.81} &
  \textbf{0.4 (0.26)} \\ \hline
\multicolumn{1}{|c|}{ADA} &
  \multicolumn{1}{c|}{\textbf{3.422 (3.08 - 3.802)}} &
  \multicolumn{1}{c|}{\textbf{0.79}} &
  \multicolumn{1}{c|}{\textbf{0.44 (0.23)}} &
  \multicolumn{1}{c|}{3.103 (2.783 - 3.46)} &
  \multicolumn{1}{c|}{0.76} &
  0.43 (0.21) \\ \hline
\multicolumn{1}{|c|}{GNB} &
  \multicolumn{1}{c|}{\textbf{2.479 (2.21 - 2.781)}} &
  \multicolumn{1}{c|}{\textbf{0.76}} &
  \multicolumn{1}{c|}{0.23 (0.13)} &
  \multicolumn{1}{c|}{2.191 (1.998 - 2.402)} &
  \multicolumn{1}{c|}{0.73} &
  \textbf{0.28 (0.18)} \\ \hline
\multicolumn{1}{|c|}{LDA} &
  \multicolumn{1}{c|}{\textbf{4.087 (3.65 - 4.577)}} &
  \multicolumn{1}{c|}{\textbf{0.84}} &
  \multicolumn{1}{c|}{\textbf{0.46 (0.3)}} &
  \multicolumn{1}{c|}{3.973 (3.509 - 4.498)} &
  \multicolumn{1}{c|}{0.81} &
  0.42 (0.26) \\ \hline
\multicolumn{1}{|c|}{QDA} &
  \multicolumn{1}{c|}{3.454 (3.094 - 3.857)} &
  \multicolumn{1}{c|}{\textbf{0.82}} &
  \multicolumn{1}{c|}{0.38 (0.25)} &
  \multicolumn{1}{c|}{\textbf{3.514 (3.133 - 3.941)}} &
  \multicolumn{1}{c|}{\textbf{0.82}} &
  \textbf{0.39 (0.26)} \\ \hline
\multicolumn{1}{|c|}{MLP} &
  \multicolumn{1}{c|}{3.341 (3.009 - 3.711)} &
  \multicolumn{1}{c|}{\textbf{0.81}} &
  \multicolumn{1}{c|}{0.39 (0.22)} &
  \multicolumn{1}{c|}{\textbf{3.532 (3.146 - 3.964)}} &
  \multicolumn{1}{c|}{0.80} &
  \textbf{0.4 (0.23)} \\ \hline
\multicolumn{1}{|c|}{XGB} &
  \multicolumn{1}{c|}{3.152 (2.833 - 3.507)} &
  \multicolumn{1}{c|}{\textbf{0.81}} &
  \multicolumn{1}{c|}{\textbf{0.34 (0.22)}} &
  \multicolumn{1}{c|}{\textbf{3.235 (2.866 - 3.652)}} &
  \multicolumn{1}{c|}{0.79} &
  \textbf{0.35 (0.22)} \\ \hline
\end{tabular}
\caption{Performance metrics, with and without the inclusion of ISMs, of the proposed method for uncertainty estimation, compared against the absolute difference of the classification probability from 0.5 for the predicted class. Metrics highlighted in bold reflect the best performance for a given method.}
\label{tab:performance}
\end{table}
\end{landscape}
\normalsize

\section{Applications for Abstention and Explainability}\label{sec:Applications}
\noindent The following section explores the application of the proposed meta-heuristics and uncertainty estimation system for abstaining from making a decision and explaining the level of uncertainty in terms of the sources of decision-making complexity captured by the proposed meta-heuristics.

\subsection{Abstention}
\begin{figure}[ht]
    \centering
    \centerline{\includegraphics[scale=0.4]{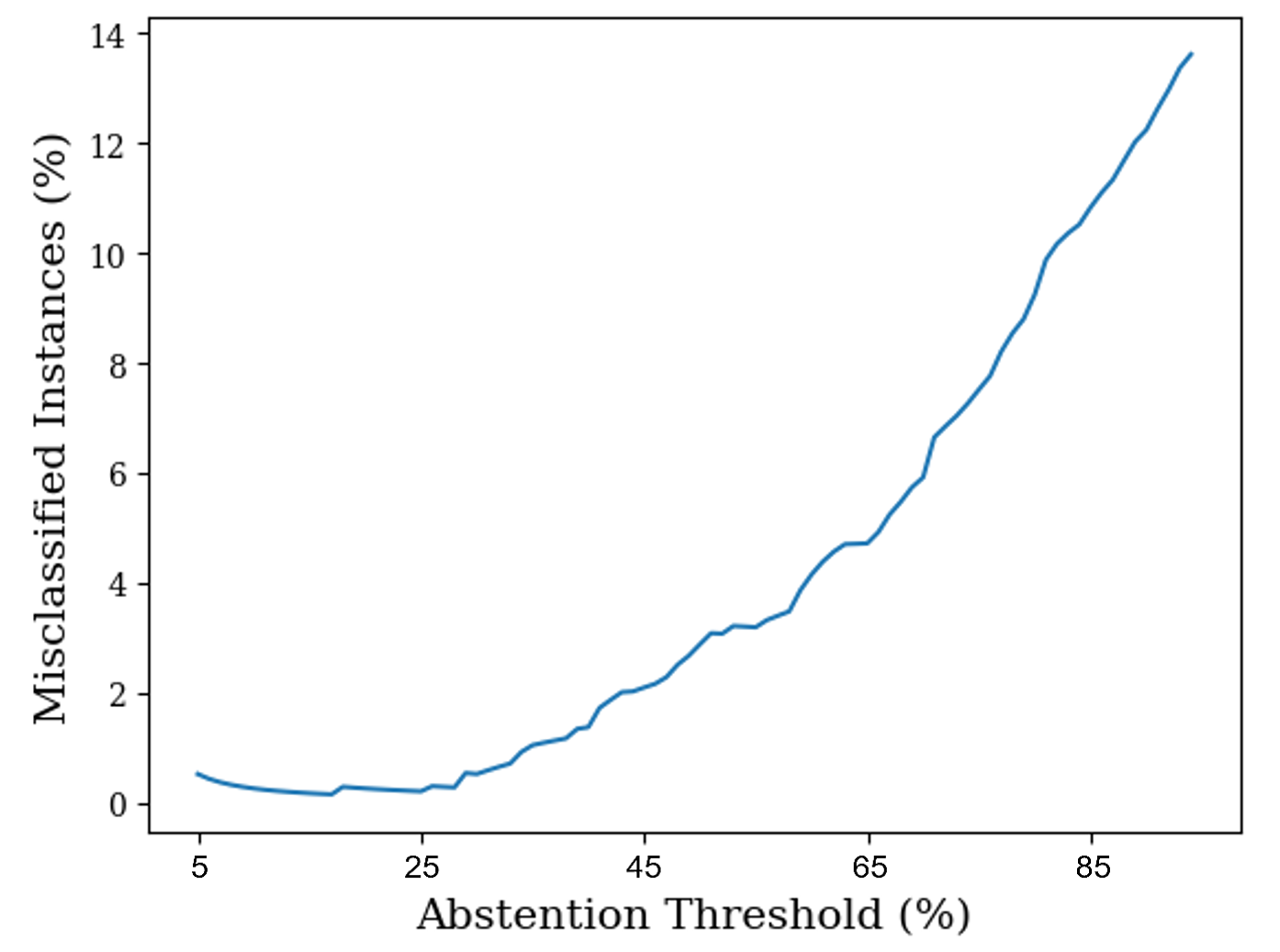}}
    \caption{Percentage of misclassified instances for a given abstention threshold.}
    \label{fig:Abstention}
\end{figure}

\noindent In terms of facilitating trust calibration, the estimated uncertainty can be used as a method for abstention and built into the deployed model, preventing decisions from being shown to the end-user when uncertainty surpasses a given threshold. To demonstrate this concept, an abstention threshold, beginning at the 5th percentile and then incrementally increasing until the 95th percentile, was applied to the uncertainty estimates for the sepsis dataset used in subsection~\ref{subsec:External_Eval}. At each threshold, the percentage of misclassified instances was calculated. Fig.~\ref{fig:Abstention} demonstrates how by varying the threshold for abstention the number of misclassifications can be reduced. Although such an approach can reduce the number of misclassification, the number of instances for which the model is applicable for also reduces. Therefore, the application of such a method may not be appropriate for all use cases.

\subsection{Explainability}
\noindent The second application of the proposed methods is for the improved explanation of predictive uncertainty. Shapley additive explanations (SHAP) can be applied to the calculated meta-heuristics and the estimated uncertainty to uncertainty to demonstrate how each concept of complexity influences the level of certainty in decision-making. As an example, SHAP values were calculated for a patient with high predictive uncertainty and a patient with low predictive uncertainty within the Sepsis-3 dataset used in section~\ref{subsec:External_Eval}, when an MLP model is applied. 

\begin{figure}[!h]
    \centering
    \centerline{\includegraphics[scale=0.7]{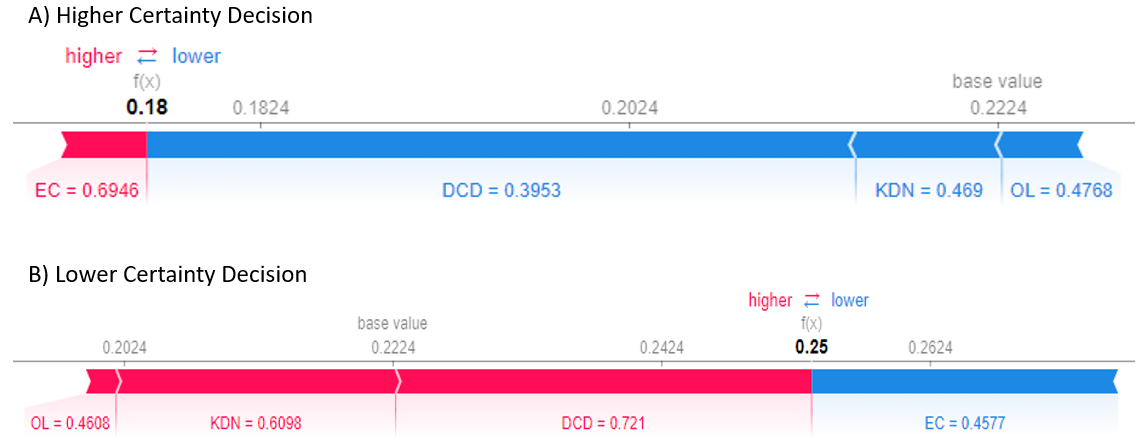}}
    \caption{Force plots demonstrating the impact of the meta-heuristics on the level of uncertainty of an MLP model for predicting 30-day mortality of sepsis-3 patients. Red banners indicate the meta-feature value is increasing the level of uncertainty of a prediction and the blue banner indicates the meta-heuristic is decreasing the level of uncertainty. The base value indicates the mean estimated uncertainty for instances in the training set and the f(x) refers to the uncertainty value estimated for the current patient.}
    \label{fig:Force plots}
\end{figure}

Fig.~\ref{fig:Force plots} shows the force plots for two patients, for patient A the classification decision was regarded as having a higher degree of certainty than the mean from the training set, and for patient B, the decision was regarded as lower certainty. The benefit of the proposed meta-heuristics is that they are a mathematical derivation of explainable concepts within humanistic decision-making and therefore can be expressed using natural language. Therefore, for patient A we know that the decision was easier primarily because there was less diversity in the class outcomes of similar patients, as evidenced by KDN and DCD, with the level of outlierness playing a small role in further reducing uncertainty. The only factor which increased the uncertainty of the decision for patient A was a higher degree of evidence supporting both class outcomes, evidenced by the EC score. Regarding patient (B), the opposite is true, both KDN and DCD are high, which indicates a higher level of diversity in class outcomes among similar patients, with the level of outlierness increasing the level of uncertainty. However, there is less conflicting evidence for patient B which works to lower the level of uncertainty. The benefit of using interpretable methods to explore AI uncertainty is that they can be used to further understand why a model may abstain from making a decision, or prompt further investigation into why patients may be classified incorrectly.

\section{Discussion}\label{sec:Discussion}
\noindent The proposed method has been demonstrated to be effective in estimating the uncertainty associated with AI decision-making, which can be a crucial factor in instilling trust in AI applications, particularly in high-stakes domains such as healthcare.There are several clear avenues for the application of such methods to improve both the robustness of AI development and the facilitation of trust-calibration. 

Section~\ref{sec:Applications} presents a clear use case for the estimated uncertainty scores to be used as a means of abstaining from making a prediction. Although the present study did not fully explore the potential of the proposed methods for model abstention, prior research has suggested methods for determining the optimal points for abstention~\cite{fischer2016optimal} and abstention-specific performance measures~\cite{condessa2017performance}. Therefore, future research in this area could consider incorporating the proposed methods for uncertainty estimation within existing frameworks for developing classification systems with the option of abstention.

To enhance the robustness of AI development, the proposed meta-heuristics can also be utilised to identify the competency of classifiers. The use of meta-information to understand the strengths and weakness of algorithms is a long-standing practice~\cite{smith2014instance,houston2022genetically,scholz2021comparison,lango2022makes}. The class-independent and highly explainable nature of the proposed meta-heuristics makes them suitable for algorithmic development, and their application in this area can offer new insights into how the algorithm may perform on unseen data.

An emerging area in the field of precision medicine is the personalisation of model development to the patient instead of the overall task. Studies have investigated meta-learning techniques for dynamic classifier selection~\cite{cruz2018dynamic} and ensemble generation~\cite{cruz2015meta, guo2021dynamic} in this domain. Within the area of complexity, such methods could be applied to select models based on their ability to handle more challenging instances to maximise the likelihood of a correct classification~\cite{cruz2017dynamic}. In domains where large datasets are scarce, meta-learning approaches have the potential to overcome data limitations and learn viable solutions to parameter tuning and model architecture from similar problems before applying the learned principles to the current problem to arrive at a more optimal solution.

Qualitative research indicates that different methods are employed when addressing uncertainty in real-world human decision-making, depending on the source of the uncertainty, with factors such as past experience and evidence availability influencing the strategy chosen~\cite{kushniruk2001analysis,yang2019unremarkable}. The advantage of the proposed methods is that they are explainable and can identify the causes of uncertainty within the AI decision-making process, enabling clinicians to reason with the output. By providing end-users with a more transparent means of engaging with and comprehending model outputs, it is hypothesized that trust calibration will be enhanced.

In conclusion, the proposed method of uncertainty estimation was effective in identifying instances that are more likely to be misclassified. Furthermore, the proposed meta-heuristics offer additional opportunities to improve the explainability of AI decision-making. Future research directions include the application of the proposed meta-heuristics to enhance model development through meta-learning and studying the impact of natural language explanations of uncertainty on the trust calibration of end-users.

\newpage
\section*{Acknowledgements}
We wish to acknowledge the financial support of DMRC and Loughborough University who jointly funded the project.

\section*{Author contributions}
A.H. and G.C. contributed to the design of the study. A.H. and G.C. conceived the experiments, A.H. conducted the experiment(s). A.H. analysed the results. A.H. and G.C. interpreted the findings. G.C. supervised the project. A.H. drafted the initial version of the manuscript. A.H. and G.C. reviewed the manuscript. 

\section*{Competing Interests Statement}
The authors declare they have no competing interests to disclose.

\newpage

\section*{Supplementary File}
\label{sec:Supp Files}
\renewcommand{\thetable}{S.\arabic{table}}
\setcounter{table}{0}

\begin{table}[h]
\centering
\begin{tabular}{|l|l|}
\hline
\textbf{Model} & \textbf{Hyperparameter}                                                                                        \\ \hline
LR             & \begin{tabular}[c]{@{}l@{}}solver:   liblinear; saga  \\ penalty: l1; l2\\ regularisation: (0.01, 100)\end{tabular}         \\ \hline
SVM & \begin{tabular}[c]{@{}l@{}}regularisation:   (0.01, 100)\\ gamma: (0.01, 100)\\ degree: (1,5)\\ kernel: linear; polynomial; rbf; sigmoid\end{tabular}                          \\ \hline
KNN            & \begin{tabular}[c]{@{}l@{}}no. of neighbors:   (2,11)\\ algorithm: auto; ball\_tree; kd\_tree; brute\end{tabular}  \\ \hline
RF             & \begin{tabular}[c]{@{}l@{}}maximum depth:   (2,10)\\ no. of estimators: (5,20)\\ maximum features: (0.25,1)\end{tabular} \\ \hline
ADA            & \begin{tabular}[c]{@{}l@{}}learning rate:   (0.005, 0.9)\\  no. of estimators: (5, 20)\end{tabular}               \\ \hline
NB             & variable smoothing   : (0.01,1)                                                                                    \\ \hline
LDA            & solver: svd; lsqr; eigen                                                                                    \\ \hline
QDA            & \begin{tabular}[c]{@{}l@{}}regularisation parameter:   (0.00001, 0.1)\\ tol: (0.0001, 0.1)\end{tabular}                      \\ \hline
MLP & \begin{tabular}[c]{@{}l@{}}activation function:   tanh; relu\\      solver: sgd; adam\\      alpha: (0.0001, 0.05)\\      learning rate: constant; adaptive\end{tabular}  \\ \hline
XGB & \begin{tabular}[c]{@{}l@{}}no. of estimators:   (5, 20)\\      maximum depth: (2, 15)\\      learning rate: (0.05, 0.20)\\      min. child weight: (1, 4)\end{tabular} \\ \hline
\end{tabular}
\caption{Hyperparameters tuned using Bayesian Cross-Validation for each ML algorithm.}
\label{tab:hyperparameters}
\end{table}


\begin{thebibliography}{10}

\bibitem{abad2021detecting}
Zahra Shakeri~Hossein Abad and Joon Lee.
\newblock Detecting uncertainty of mortality prediction using confident
  learning.
\newblock In {\em 2021 43rd Annual International Conference of the IEEE
  Engineering in Medicine \& Biology Society (EMBC)}, pages 1719--1722. IEEE,
  2021.

\bibitem{acuna2005empirical}
Edgar Acu{\~n}a and Caroline Rodr{\'\i}guez.
\newblock An empirical study of the effect of outliers on the misclassification
  error rate.
\newblock {\em Submitted to Transactions on Knowledge and Data Engineering},
  2005.

\bibitem{amann2020explainability}
Julia Amann, Alessandro Blasimme, Effy Vayena, Dietmar Frey, Vince~I Madai, and
  Precise4Q Consortium.
\newblock Explainability for artificial intelligence in healthcare: a
  multidisciplinary perspective.
\newblock {\em BMC medical informatics and decision making}, 20:1--9, 2020.

\bibitem{antal2014ensemble}
B{\'a}lint Antal and Andr{\'a}s Hajdu.
\newblock An ensemble-based system for automatic screening of diabetic
  retinopathy.
\newblock {\em Knowledge-based systems}, 60:20--27, 2014.

\bibitem{asan2020artificial}
Onur Asan, Alparslan~Emrah Bayrak, Avishek Choudhury, et~al.
\newblock Artificial intelligence and human trust in healthcare: focus on
  clinicians.
\newblock {\em Journal of medical Internet research}, 22(6):e15154, 2020.

\bibitem{barandas2022uncertainty}
Mar{\'\i}lia Barandas, Duarte Folgado, Ricardo Santos, Raquel Sim{\~a}o, and
  Hugo Gamboa.
\newblock Uncertainty-based rejection in machine learning: Implications for
  model development and interpretability.
\newblock {\em Electronics}, 11(3):396, 2022.

\bibitem{buolamwini2017gender}
Joy~Adowaa Buolamwini.
\newblock {\em Gender shades: intersectional phenotypic and demographic
  evaluation of face datasets and gender classifiers}.
\newblock PhD thesis, Massachusetts Institute of Technology, 2017.

\bibitem{chang2003statistical}
Edward~Y Chang, Beitao Li, Gang Wu, and Kingshy Goh.
\newblock Statistical learning for effective visual information retrieval.
\newblock In {\em Proceedings 2003 International Conference on Image Processing
  (Cat. No. 03CH37429)}, volume~3, pages III--609. IEEE, 2003.

\bibitem{chawla2002smote}
Nitesh~V Chawla, Kevin~W Bowyer, Lawrence~O Hall, and W~Philip Kegelmeyer.
\newblock Smote: synthetic minority over-sampling technique.
\newblock {\em Journal of artificial intelligence research}, 16:321--357, 2002.

\bibitem{chicco2020machine}
Davide Chicco and Giuseppe Jurman.
\newblock Machine learning can predict survival of patients with heart failure
  from serum creatinine and ejection fraction alone.
\newblock {\em BMC medical informatics and decision making}, 20(1):1--16, 2020.

\bibitem{cioffi2001study}
Jane Cioffi.
\newblock A study of the use of past experiences in clinical decision making in
  emergency situations.
\newblock {\em International journal of nursing studies}, 38(5):591--599, 2001.

\bibitem{cole2014impact}
Elodia~B Cole, Zheng Zhang, Helga~S Marques, R~Edward Hendrick, Martin~J Yaffe,
  and Etta~D Pisano.
\newblock Impact of computer-aided detection systems on radiologist accuracy
  with digital mammography.
\newblock {\em AJR. American journal of roentgenology}, 203(4):909, 2014.

\bibitem{condessa2017performance}
Filipe Condessa, Jos{\'e} Bioucas-Dias, and Jelena Kova{\v{c}}evi{\'c}.
\newblock Performance measures for classification systems with rejection.
\newblock {\em Pattern Recognition}, 63:437--450, 2017.

\bibitem{cruz2018dynamic}
Rafael~MO Cruz, Robert Sabourin, and George~DC Cavalcanti.
\newblock Dynamic classifier selection: Recent advances and perspectives.
\newblock {\em Information Fusion}, 41:195--216, 2018.

\bibitem{cruz2015meta}
Rafael~MO Cruz, Robert Sabourin, George~DC Cavalcanti, and Tsang~Ing Ren.
\newblock Meta-des: A dynamic ensemble selection framework using meta-learning.
\newblock {\em Pattern recognition}, 48(5):1925--1935, 2015.

\bibitem{cruz2017dynamic}
Rafael~MO Cruz, Hiba~H Zakane, Robert Sabourin, and George~DC Cavalcanti.
\newblock Dynamic ensemble selection vs k-nn: why and when dynamic selection
  obtains higher classification performance?
\newblock In {\em 2017 Seventh International Conference on Image Processing
  Theory, Tools and Applications (IPTA)}, pages 1--6. IEEE, 2017.

\bibitem{czerniak2003application}
Jacek Czerniak and Hubert Zarzycki.
\newblock Application of rough sets in the presumptive diagnosis of urinary
  system diseases.
\newblock In {\em Artificial intelligence and security in computing systems},
  pages 41--51. Springer, 2003.

\bibitem{dallora2017machine}
Ana~Luiza Dallora, Shahryar Eivazzadeh, Emilia Mendes, Johan Berglund, and
  Peter Anderberg.
\newblock Machine learning and microsimulation techniques on the prognosis of
  dementia: A systematic literature review.
\newblock {\em PloS one}, 12(6):e0179804, 2017.

\bibitem{debernardi2020combination}
Silvana Debernardi, Harrison O’Brien, Asma~S Algahmdi, Nuria Malats, Grant~D
  Stewart, Marija Plje{\v{s}}a-Ercegovac, Eithne Costello, William Greenhalf,
  Amina Saad, Rhiannon Roberts, et~al.
\newblock A combination of urinary biomarker panel and pancrisk score for
  earlier detection of pancreatic cancer: A case--control study.
\newblock {\em PLoS Medicine}, 17(12):e1003489, 2020.

\bibitem{Dua:2019}
Dheeru Dua and Casey Graff.
\newblock {UCI} machine learning repository, 2019.

\bibitem{elkan2001foundations}
Charles Elkan.
\newblock The foundations of cost-sensitive learning.
\newblock In {\em International joint conference on artificial intelligence},
  volume~17, pages 973--978. Lawrence Erlbaum Associates Ltd, 2001.

\bibitem{fischer2016optimal}
Lydia Fischer, Barbara Hammer, and Heiko Wersing.
\newblock Optimal local rejection for classifiers.
\newblock {\em Neurocomputing}, 214:445--457, 2016.

\bibitem{ghosh2019imbalanced}
Kushankur Ghosh, Arghasree Banerjee, Sankhadeep Chatterjee, and Soumya Sen.
\newblock Imbalanced twitter sentiment analysis using minority oversampling.
\newblock In {\em 2019 IEEE 10th international conference on awareness science
  and technology (iCAST)}, pages 1--5. IEEE, 2019.

\bibitem{gil2012predicting}
David Gil, Jose~Luis Girela, Joaquin De~Juan, M~Jose Gomez-Torres, and Magnus
  Johnsson.
\newblock Predicting seminal quality with artificial intelligence methods.
\newblock {\em Expert Systems with Applications}, 39(16):12564--12573, 2012.

\bibitem{golovenkin2020trajectories}
Sergey~E Golovenkin, Jonathan Bac, Alexander Chervov, Evgeny~M Mirkes, Yuliya~V
  Orlova, Emmanuel Barillot, Alexander~N Gorban, and Andrei Zinovyev.
\newblock Trajectories, bifurcations, and pseudo-time in large clinical
  datasets: Applications to myocardial infarction and diabetes data.
\newblock {\em GigaScience}, 9(11):giaa128, 2020.

\bibitem{guo2021dynamic}
Chonghui Guo, Mucan Liu, and Menglin Lu.
\newblock A dynamic ensemble learning algorithm based on k-means for icu
  mortality prediction.
\newblock {\em Applied Soft Computing}, 103:107166, 2021.

\bibitem{ho2002complexity}
Tin~Kam Ho and Mitra Basu.
\newblock Complexity measures of supervised classification problems.
\newblock {\em IEEE transactions on pattern analysis and machine intelligence},
  24(3):289--300, 2002.

\bibitem{hou2020predicting}
Nianzong Hou, Mingzhe Li, Lu~He, Bing Xie, Lin Wang, Rumin Zhang, Yong Yu,
  Xiaodong Sun, Zhengsheng Pan, and Kai Wang.
\newblock Predicting 30-days mortality for mimic-iii patients with sepsis-3: a
  machine learning approach using xgboost.
\newblock {\em Journal of translational medicine}, 18(1):1--14, 2020.

\bibitem{houston2022genetically}
Andrew Houston and Georgina Cosma.
\newblock A genetically-optimised artificial life algorithm for
  complexity-based synthetic dataset generation.
\newblock {\em Information Sciences}, 2022.

\bibitem{houston2021predicting}
Andrew Houston, Georgina Cosma, Phillipa Turner, and Alexander Bennett.
\newblock Predicting surgical outcomes for chronic exertional compartment
  syndrome using a machine learning framework with embedded trust by
  interrogation strategies.
\newblock {\em Scientific Reports}, 11, 2021.

\bibitem{hullermeier2021aleatoric}
Eyke H{\"u}llermeier and Willem Waegeman.
\newblock Aleatoric and epistemic uncertainty in machine learning: An
  introduction to concepts and methods.
\newblock {\em Machine Learning}, 110(3):457--506, 2021.

\bibitem{islam2020likelihood}
MM~Islam, Rahatara Ferdousi, Sadikur Rahman, and Humayra~Yasmin Bushra.
\newblock Likelihood prediction of diabetes at early stage using data mining
  techniques.
\newblock In {\em Computer Vision and Machine Intelligence in Medical Image
  Analysis}, pages 113--125. Springer, 2020.

\bibitem{islam2014heuristics}
Roosan Islam, Charlene Weir, and Guilherme Del~Fiol.
\newblock Heuristics in managing complex clinical decision tasks in experts'
  decision making.
\newblock In {\em 2014 IEEE International Conference on Healthcare
  Informatics}, pages 186--193. IEEE, 2014.

\bibitem{karhade2019development}
Aditya~V Karhade, Quirina~CBS Thio, Paul~T Ogink, Akash~A Shah, Christopher~M
  Bono, Kevin~S Oh, Phil~J Saylor, Andrew~J Schoenfeld, John~H Shin, Mitchel~B
  Harris, et~al.
\newblock Development of machine learning algorithms for prediction of 30-day
  mortality after surgery for spinal metastasis.
\newblock {\em Neurosurgery}, 85(1):E83--E91, 2019.

\bibitem{kaur2022sensible}
Harmanpreet Kaur, Eytan Adar, Eric Gilbert, and Cliff Lampe.
\newblock Sensible ai: Re-imagining interpretability and explainability using
  sensemaking theory.
\newblock In {\em 2022 ACM Conference on Fairness, Accountability, and
  Transparency}, pages 702--714, 2022.

\bibitem{kohli2018cad}
Ajay Kohli and Saurabh Jha.
\newblock Why cad failed in mammography.
\newblock {\em Journal of the American College of Radiology}, 15(3):535--537,
  2018.

\bibitem{kompa2021second}
Benjamin Kompa, Jasper Snoek, and Andrew~L Beam.
\newblock Second opinion needed: communicating uncertainty in medical machine
  learning.
\newblock {\em NPJ Digital Medicine}, 4(1):4, 2021.

\bibitem{kourou2015machine}
Konstantina Kourou, Themis~P Exarchos, Konstantinos~P Exarchos, Michalis~V
  Karamouzis, and Dimitrios~I Fotiadis.
\newblock Machine learning applications in cancer prognosis and prediction.
\newblock {\em Computational and structural biotechnology journal}, 13:8--17,
  2015.

\bibitem{kushniruk2001analysis}
Andre~W Kushniruk.
\newblock Analysis of complex decision-making processes in health care:
  cognitive approaches to health informatics.
\newblock {\em Journal of biomedical informatics}, 34(5):365--376, 2001.

\bibitem{lamari2021smote}
Mouna Lamari, Nabiha Azizi, Nacer~Eddine Hammami, Assia Boukhamla, Soraya
  Cheriguene, Najdette Dendani, and Nacer~Eddine Benzebouchi.
\newblock Smote--enn-based data sampling and improved dynamic ensemble
  selection for imbalanced medical data classification.
\newblock In {\em Advances on Smart and Soft Computing: Proceedings of ICACIn
  2020}, pages 37--49. Springer, 2021.

\bibitem{lango2022makes}
Mateusz Lango and Jerzy Stefanowski.
\newblock What makes multi-class imbalanced problems difficult? an experimental
  study.
\newblock {\em Expert Systems with Applications}, 199:116962, 2022.

\bibitem{lewis1994sequential}
David~D. Lewis and William~A. Gale.
\newblock A sequential algorithm for training text classifiers.
\newblock In Bruce~W. Croft and C.~J. van Rijsbergen, editors, {\em SIGIR '94},
  pages 3--12, London, 1994. Springer London.

\bibitem{little2008suitability}
Max Little, Patrick McSharry, Eric Hunter, Jennifer Spielman, and Lorraine
  Ramig.
\newblock Suitability of dysphonia measurements for telemonitoring of
  parkinson’s disease.
\newblock {\em Nature Precedings}, pages 1--1, 2008.

\bibitem{lopez2013insight}
Victoria L{\'o}pez, Alberto Fern{\'a}ndez, Salvador Garc{\'\i}a, Vasile Palade,
  and Francisco Herrera.
\newblock An insight into classification with imbalanced data: Empirical
  results and current trends on using data intrinsic characteristics.
\newblock {\em Information sciences}, 250:113--141, 2013.

\bibitem{lundberg2017unified}
Scott~M Lundberg and Su-In Lee.
\newblock A unified approach to interpreting model predictions.
\newblock {\em Advances in neural information processing systems}, 30, 2017.

\bibitem{macia2014towards}
Nuria Macia and Ester Bernad{\'o}-Mansilla.
\newblock Towards uci+: a mindful repository design.
\newblock {\em Information Sciences}, 261:237--262, 2014.

\bibitem{mckinney2020international}
Scott~Mayer McKinney, Marcin Sieniek, Varun Godbole, Jonathan Godwin, Natasha
  Antropova, Hutan Ashrafian, Trevor Back, Mary Chesus, Greg~S Corrado, Ara
  Darzi, et~al.
\newblock International evaluation of an ai system for breast cancer screening.
\newblock {\em Nature}, 577(7788):89--94, 2020.

\bibitem{northcutt2021confident}
Curtis Northcutt, Lu~Jiang, and Isaac Chuang.
\newblock Confident learning: Estimating uncertainty in dataset labels.
\newblock {\em Journal of Artificial Intelligence Research}, 70:1373--1411,
  2021.

\bibitem{ortigosa2017measuring}
Jonathan Ortigosa-Hern{\'a}ndez, Inaki Inza, and Jose~A Lozano.
\newblock Measuring the class-imbalance extent of multi-class problems.
\newblock {\em Pattern Recognition Letters}, 98:32--38, 2017.

\bibitem{patel2021artificial}
Urvish~K Patel, Arsalan Anwar, Sidra Saleem, Preeti Malik, Bakhtiar Rasul,
  Karan Patel, Robert Yao, Ashok Seshadri, Mohammed Yousufuddin, and
  Kogulavadanan Arumaithurai.
\newblock Artificial intelligence as an emerging technology in the current care
  of neurological disorders.
\newblock {\em Journal of neurology}, 268(5):1623--1642, 2021.

\bibitem{patricio2018using}
Miguel Patr{\'\i}cio, Jos{\'e} Pereira, Joana Cris{\'o}stomo, Paulo Matafome,
  Manuel Gomes, Raquel Sei{\c{c}}a, and Francisco Caramelo.
\newblock Using resistin, glucose, age and bmi to predict the presence of
  breast cancer.
\newblock {\em BMC cancer}, 18(1):1--8, 2018.

\bibitem{platt1999probabilistic}
John Platt.
\newblock Probabilistic outputs for support vector machines and comparisons to
  regularized likelihood methods.
\newblock {\em Advances in large margin classifiers}, 10(3):61--74, 1999.

\bibitem{polce2020development}
Evan~M Polce, Kyle~N Kunze, Michael Fu, Grant~E Garrigues, Brian Forsythe,
  Gregory~P Nicholson, Brian~J Cole, and Nikhil~N Verma.
\newblock Development of supervised machine learning algorithms for prediction
  of satisfaction at two years following total shoulder arthroplasty.
\newblock {\em Journal of Shoulder and Elbow Surgery}, 2020.

\bibitem{prati2015class}
Ronaldo~C Prati, Gustavo~EAPA Batista, and Diego~F Silva.
\newblock Class imbalance revisited: a new experimental setup to assess the
  performance of treatment methods.
\newblock {\em Knowledge and Information Systems}, 45(1):247--270, 2015.

\bibitem{quinlan1986induction}
J.~Ross Quinlan.
\newblock Induction of decision trees.
\newblock {\em Machine learning}, 1:81--106, 1986.

\bibitem{rechkemmer2022confidence}
Amy Rechkemmer and Ming Yin.
\newblock When confidence meets accuracy: Exploring the effects of multiple
  performance indicators on trust in machine learning models.
\newblock In {\em Proceedings of the 2022 chi conference on human factors in
  computing systems}, pages 1--14, 2022.

\bibitem{ribeiro2016should}
Marco~Tulio Ribeiro, Sameer Singh, and Carlos Guestrin.
\newblock " why should i trust you?" explaining the predictions of any
  classifier.
\newblock In {\em Proceedings of the 22nd ACM SIGKDD international conference
  on knowledge discovery and data mining}, pages 1135--1144, 2016.

\bibitem{roy2001toward}
Nicholas Roy and Andrew McCallum.
\newblock Toward optimal active learning through sampling estimation of error
  reduction.
\newblock In {\em International Conference on Machine Learning}, pages
  441–--448. Morgan Kaufmann, 2001.

\bibitem{sakar2019comparative}
C~Okan Sakar, Gorkem Serbes, Aysegul Gunduz, Hunkar~C Tunc, Hatice Nizam,
  Betul~Erdogdu Sakar, Melih Tutuncu, Tarkan Aydin, M~Erdem Isenkul, and Hulya
  Apaydin.
\newblock A comparative analysis of speech signal processing algorithms for
  parkinson’s disease classification and the use of the tunable q-factor
  wavelet transform.
\newblock {\em Applied Soft Computing}, 74:255--263, 2019.

\bibitem{scholz2021comparison}
Michael Scholz and Tristan Wimmer.
\newblock A comparison of classification methods across different data
  complexity scenarios and datasets.
\newblock {\em Expert Systems with Applications}, 168:114217, 2021.

\bibitem{sharma2017evidence}
Manali Sharma and Mustafa Bilgic.
\newblock Evidence-based uncertainty sampling for active learning.
\newblock {\em Data Mining and Knowledge Discovery}, 31:164--202, 2017.

\bibitem{smith2011improving}
Michael~R Smith and Tony Martinez.
\newblock Improving classification accuracy by identifying and removing
  instances that should be misclassified.
\newblock In {\em The 2011 International Joint Conference on Neural Networks},
  pages 2690--2697. IEEE, 2011.

\bibitem{smith2014instance}
Michael~R Smith, Tony Martinez, and Christophe Giraud-Carrier.
\newblock An instance level analysis of data complexity.
\newblock {\em Machine learning}, 95:225--256, 2014.

\bibitem{spiegelhalter2008understanding}
David~J Spiegelhalter.
\newblock Understanding uncertainty.
\newblock {\em The Annals of Family Medicine}, 6(3):196--197, 2008.

\bibitem{stefanowski2013overlapping}
Jerzy Stefanowski.
\newblock Overlapping, rare examples and class decomposition in learning
  classifiers from imbalanced data.
\newblock {\em Emerging paradigms in machine learning}, pages 277--306, 2013.

\bibitem{tang2017local}
Bo~Tang and Haibo He.
\newblock A local density-based approach for outlier detection.
\newblock {\em Neurocomputing}, 241:171--180, 2017.

\bibitem{taruscio2021multifactorial}
Domenica Taruscio and Alberto Mantovani.
\newblock Multifactorial rare diseases: Can uncertainty analysis bring added
  value to the search for risk factors and etiopathogenesis?
\newblock {\em Medicina}, 57(2):119, 2021.

\bibitem{tonekaboni2019clinicians}
Sana Tonekaboni, Shalmali Joshi, Melissa~D McCradden, and Anna Goldenberg.
\newblock What clinicians want: contextualizing explainable machine learning
  for clinical end use.
\newblock In {\em Machine learning for healthcare conference}, pages 359--380.
  PMLR, 2019.

\bibitem{tsanas2013objective}
Athanasios Tsanas, Max~A Little, Cynthia Fox, and Lorraine~O Ramig.
\newblock Objective automatic assessment of rehabilitative speech treatment in
  parkinson's disease.
\newblock {\em IEEE Transactions on Neural Systems and Rehabilitation
  Engineering}, 22(1):181--190, 2013.

\bibitem{van2002gene}
Marc~J Van De~Vijver, Yudong~D He, Laura~J Van't~Veer, Hongyue Dai,
  Augustinus~AM Hart, Dorien~W Voskuil, George~J Schreiber, Johannes~L Peterse,
  Chris Roberts, Matthew~J Marton, et~al.
\newblock A gene-expression signature as a predictor of survival in breast
  cancer.
\newblock {\em New England Journal of Medicine}, 347(25):1999--2009, 2002.

\bibitem{vuttipittayamongkol2021class}
Pattaramon Vuttipittayamongkol, Eyad Elyan, and Andrei Petrovski.
\newblock On the class overlap problem in imbalanced data classification.
\newblock {\em Knowledge-based systems}, 212:106631, 2021.

\bibitem{xie2020chexplain}
Yao Xie, Melody Chen, David Kao, Ge~Gao, and Xiang'Anthony' Chen.
\newblock Chexplain: enabling physicians to explore and understand data-driven,
  ai-enabled medical imaging analysis.
\newblock In {\em Proceedings of the 2020 CHI Conference on Human Factors in
  Computing Systems}, pages 1--13, 2020.

\bibitem{yang2019unremarkable}
Qian Yang, Aaron Steinfeld, and John Zimmerman.
\newblock Unremarkable ai: Fitting intelligent decision support into critical,
  clinical decision-making processes.
\newblock In {\em Proceedings of the 2019 CHI conference on human factors in
  computing systems}, pages 1--11, 2019.

\bibitem{zhu2018lrid}
Rui Zhu, Ziyu Wang, Zhanyu Ma, Guijin Wang, and Jing-Hao Xue.
\newblock Lrid: A new metric of multi-class imbalance degree based on
  likelihood-ratio test.
\newblock {\em Pattern Recognition Letters}, 116:36--42, 2018.

\bibitem{zikeba2014boosted}
Maciej Zi\c{e}ba, Jakub~M Tomczak, Marek Lubicz, and Jerzy Świ\c{a}tek.
\newblock Boosted svm for extracting rules from imbalanced data in application
  to prediction of the post-operative life expectancy in the lung cancer
  patients.
\newblock {\em Applied soft computing}, 14:99--108, 2014.

\end{thebibliography}


\newpage
\end{document}